\documentclass{article}
\usepackage{iclr2025_conference,times}

\iclrfinalcopy

\usepackage[utf8]{inputenc}
\usepackage[T1]{fontenc}
\usepackage{booktabs}
\usepackage{amsmath,amsfonts,amssymb,mathtools}
\usepackage{url}
\usepackage{bbold}
\usepackage{wrapfig}
\usepackage{graphicx}
\usepackage{caption}
\usepackage{subcaption}
\usepackage{tabularx}
\usepackage{float}
\usepackage[table]{xcolor}
\usepackage{multirow}
\usepackage{threeparttable}
\usepackage{enumitem}
\usepackage{dsfont}
\usepackage{bm}
\usepackage[ruled,vlined,linesnumbered]{algorithm2e}
\usepackage{hyperref}

\bibpunct[, ]{(}{)}{;}{a}{}{,}
\renewcommand\bibfont{\fontsize{10}{12}\selectfont}

\graphicspath{{figures/}}

\title{Generating Logically Consistent Synthetic Supply Chain Data with LLM-Driven Knowledge Graph Reasoning}

\author{
Yunbo Long\(^{1}\), Ge Zheng\(^{1}\), Liming Xu\(^{1}\), and Alexandra Brintrup\(^{1,2}\) \\
\(^{1}\)Department of Engineering, University of Cambridge, Cambridge, United Kingdom \\
\(^{2}\)The Alan Turing Institute, London, United Kingdom \\
\texttt{\{yl892, gz277, lx249, ab702\}@cam.ac.uk}
}

\begin{document}

\maketitle


\begin{abstract}
Synthetic data offers a promising solution to two persistent barriers in supply chain analytics: data scarcity and data privacy. However, for synthetic data to support operational simulation and decision-making, it must do more than reproduce the statistical distributions of real records, and also preserve the \emph{operational logic} that governs supply chain processes, including the temporal orderings, mathematical dependencies, hierarchical taxonomies, and conditional rules that make a record operationally plausible. We consider this logic as the ``physics'' of supply chain data. Existing tabular generative models are primarily optimized for distributional fidelity and downstream predictive utility, and therefore often generate records that appear statistically realistic but violate fundamental operational constraints.
This paper introduces \textbf{\textit{TabKG}}, a knowledge-graph-guided framework for logically consistent synthetic supply chain tabular data generation. TabKG constructs a \textbf{\textit{Column Relationship Knowledge Graph (CR-KG)}} to represent data operational dependencies. It uses a multi-LLM ensemble with majority voting to propose candidate relationships from column metadata, validates these relationships against real data to remove hallucinated or unsupported edges, and then uses the validated CR-KG to guide generation. Specifically, TabKG compresses the original table into independent columns, generates these columns using a latent diffusion model, and deterministically reconstructs dependent columns according to the validated relationships, enforcing logical consistency by construction with respect to the discovered operational rules.
Across two industrial supply chain datasets and two downstream classification tasks, namely late-delivery risk prediction and procurement-status classification, TabKG achieves an F1 score of up to 0.97 for operational logic reasoning. This is substantially higher than prompt-only baselines, whose F1 scores range from 0.27 to 0.55. For tabular data generation, TabKG performs comparably to four state-of-the-art tabular generative baselines in terms of fidelity, utility, and privacy, while achieving stronger logical consistency. These results show that trustworthy synthetic supply chain data should preserve not only statistical realism, but also the operational rules that govern the real system. Overall, TabKG provides a foundation for moving synthetic data beyond machine learning augmentation toward high-fidelity supply chain simulation and operational decision support. The code is available at
\href{https://github.com/Yunbo-max/TabKG}
{\textcolor{red}{https://github.com/Yunbo-max/TabKG}}.
\end{abstract}


\section{Introduction}
\label{sec:introduction}

In production research, synthetic data has been widely applied to enhance machine learning tasks critical to supply chain operations \citep{long2025leveraging}. As highlighted by recent systematic reviews \citep{vlachos2025machine}, machine learning adoption in supply chain management has grown rapidly, yet data scarcity remains a persistent barrier. Synthetic data serves as an effective alternative for data sharing across organizational boundaries and as a data augmentation technique to improve predictive performance in applications such as delivery delay prediction \citep{kong2026hierarchical}, demand forecasting \citep{arora2025forecasting}, predictive maintenance \citep{schneckenreither2021order}, and supply chain risk prediction \citep{wyrembek2025causal}. By supplementing limited real-world datasets, synthetic data enables more robust model training and better generalization in these operational contexts \citep{jordon2022synthetic}.

Recent advances in generative artificial intelligence, including large language models (LLMs) and diffusion models \citep{ho2020denoising, brown2020language}, have substantially improved the ability of generative models to approximate complex data distributions. However, for synthetic data to support supply chain simulation and operational decision-making, statistical similarity alone is insufficient. Generated records must also respect the operational logic that governs real supply chain processes, including the temporal orderings, mathematical dependencies, hierarchical taxonomies, and conditional rules that make a record operationally plausible. We consider this set of domain-specific rules as the ``physics'' of supply chain data: not physical laws in a literal sense, but the structured operational constraints by which the data is produced. This perspective aligns with the broader movement toward \textit{world models}, where generative systems are expected not only to reproduce observations, but also to capture the rules and laws that govern plausible outcomes in a domain \citep{del2025world, kusiak2020convolutional}.

\begin{figure*}[t]
    \centering
    \includegraphics[width=\textwidth]{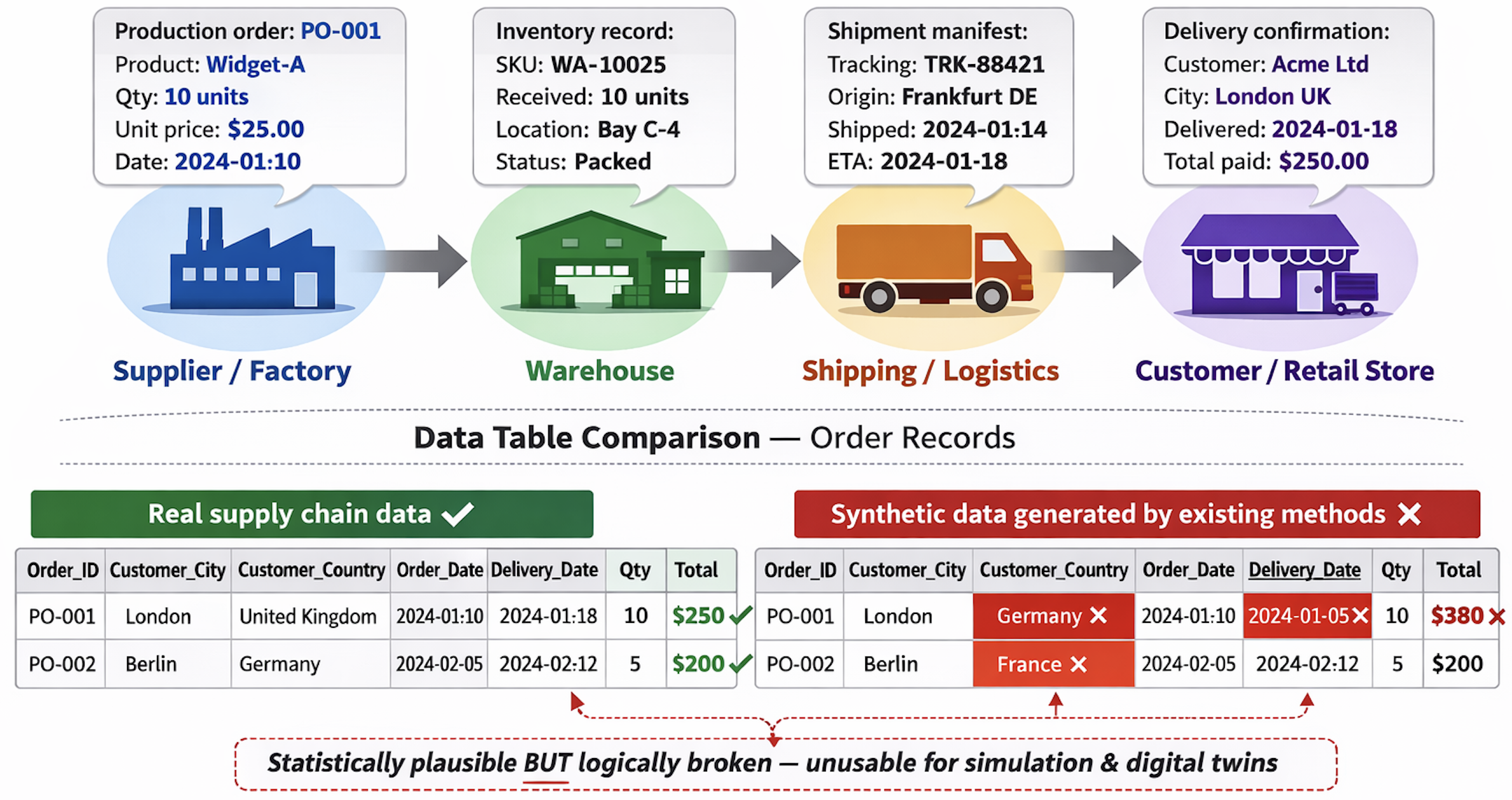}
    \caption{Illustration of inter-column logical consistency problems in synthetic supply chain data. Real data (left) maintains valid hierarchical, temporal, and mathematical relationships, while synthetic data generated by existing methods (right) frequently violates these constraints, producing records in which cities map to incorrect countries, delivery dates precede order dates, and totals do not equal price multiplied by quantity.}
    \label{fig:problem}
\end{figure*}

In supply chain management, this requirement marks a shift from generating statistically consistent data for predictive model augmentation to generating \textit{logically consistent} data that can be used in operational simulators and decision-support systems. For example, synthetic time-series orders, inventory transactions, procurement records, and delivery events may be used to drive scenario analysis, disruption planning, and operational simulation \citep{luo2022data, xu2025multi}. In such settings, a record that matches the marginal distribution of the real data may still be invalid if it violates basic process logic. Analogous to generative image models that may produce visually realistic but semantically impossible scenes, such as a car floating in the sky \citep{ho2020denoising}, a tabular generator for supply chains may produce records where delivery timestamps precede order placement, total prices do not equal unit price multiplied by quantity, or cities are mapped to incorrect countries. These inconsistencies, illustrated in \autoref{fig:problem}, make synthetic data unsuitable for realistic operational simulation \citep{polini2020digital, afif2025computer} and high-stakes decision-making.

Capturing this operational logic is particularly challenging for supply chain tabular data because the relevant constraints are neither universal nor fixed across \emph{schemas}, where a schema refers to the column-level structure of an operational table, including the names, types, and meanings of its columns. Unlike physical domains where certain constraints are broadly universal, supply chain logic varies across sectors, firms, and tables. A retail schema may encode geographical hierarchies and pricing formulas; a procurement schema may encode supplier--material mappings, order-status dependencies, and lead-time constraints; and a manufacturing schema may encode bill-of-materials relationships and production-routing logic \citep{vieira2020supply, luo2022data}. This heterogeneity means that a generative model cannot rely on a single hard-coded rule set. Instead, it must discover and enforce schema-specific operational logic automatically.

The challenge is further amplified by dimensionality and mixed data types. Real supply chain tables often contain dozens of categorical, numerical, and temporal columns connected by interleaved hierarchical, temporal, mathematical, and semantic relationships. The resulting space of possible inter-column dependencies is combinatorial, making it difficult to identify which relationships are genuine operational rules and which are merely statistical correlations. Knowledge graphs (KGs) provide a natural representation for such heterogeneous typed relationships. Prior work has shown that KGs are effective for representing relationships among supply chain entities such as suppliers, buyers, products, and facilities \citep{kosasih2024knowledge, kosasih2025trustworthy}. However, KGs have not yet been used to represent and enforce the \textit{inter-column logic of supply chain tabular data}, even though this logic has the same heterogeneous, typed-relational structure that KGs are designed to capture.

Existing deep generative models for tabular data, including GAN-based approaches such as CTGAN \citep{xu2019modeling}, diffusion-based models such as TabDDPM \citep{kotelnikov2023tabddpm} and TabSyn \citep{zhang2023mixed}, and LLM-based methods such as GReaT \citep{borisov2022language}, have demonstrated strong performance in generating statistically plausible data for machine learning tasks. Nevertheless, these models are primarily optimized to preserve marginal distributions, correlations, and downstream predictive utility, rather than explicit operational constraints. As a result, they may generate records that appear realistic under conventional fidelity metrics but violate domain-specific relationships that are essential for simulation and decision support. This reveals a critical gap: supply chain synthetic data generation requires not only distributional fidelity, but also explicit discovery and enforcement of operational logic.

To address this gap, this paper introduces \textbf{\textit{TabKG}}, a knowledge-graph-guided framework for logically consistent synthetic supply chain tabular data generation. TabKG first constructs a \textbf{\textit{Column Relationship Knowledge Graph (CR-KG)}} to represent operational dependencies among columns. It uses a multi-LLM ensemble with majority voting to propose candidate inter-column relationships from column metadata, and then validates each candidate relationship against the real data to remove hallucinated or unsupported edges. The validated CR-KG is then used to guide generation: TabKG compresses the table into independent columns, generates these columns using a latent diffusion model, and deterministically reconstructs dependent columns according to the validated relationships. In this way, logical consistency is enforced by construction rather than only encouraged through model training.
This paper makes the following contributions:
\begin{itemize}
    \item We introduce \textbf{\textit{TabKG}}, a framework that automatically discovers supply chain operational logic from tabular schemas by constructing a \textbf{\textit{Column Relationship Knowledge Graph (CR-KG)}}. TabKG combines a multi-LLM consensus mechanism with data-driven validation, leveraging semantic knowledge from LLMs while empirically pruning hallucinated or unsupported relationships without manual rule engineering.

    \item We propose a \textbf{\textit{CR-KG-guided compression-and-reconstruction generation pipeline}} that generates independent columns using latent diffusion and reconstructs dependent columns deterministically, thereby enforcing validated operational rules in the synthetic data.

    \item Across two industrial supply chain datasets and two downstream classification tasks (\emph{late-delivery risk prediction} and \emph{procurement-status classification}), TabKG achieves an F1 score of up to 0.97 for relationship discovery, substantially outperforming prompt-only baselines whose F1 scores lie in the range 0.27--0.55. It also achieves comparable performance to four state-of-the-art generative baselines on fidelity, utility, and privacy, while substantially improving logical consistency.
\end{itemize}

To the best of our knowledge, TabKG is the first method to automatically construct and enforce a knowledge-graph representation of inter-column supply chain logic for synthetic tabular data generation. Conceptually, we treat the validated CR-KG as a concrete, dataset-specific representation of the ``physics'' of supply chain operations: the temporal orderings, mathematical formulas, hierarchical taxonomies, and conditional rules that any plausible record must satisfy. Whereas existing generative models attempt to capture this operational physics implicitly through correlations in the training data, TabKG makes it explicit, validates it against real data, and uses it to constrain synthetic data generation by construction. By bridging knowledge-graph representation with generative modelling, TabKG provides a foundation for trustworthy synthetic datasets that support not only machine learning augmentation, but also high-fidelity supply chain simulation and operational decision support.

\section{Related Work}
\label{sec:related_work}
We position our contribution against three streams of prior work: (i) the characteristics of tabular data generation in supply chains that make generative modelling particularly difficult; (ii) existing AI-driven data generation and reasoning methods in production research; and (iii) the emerging push towards logically consistent generation, within which our knowledge graph-guided approach sits.

\subsection{Synthetic Data Generation in Supply Chain}
Tabular data is the dominant data modality in production and supply chain management \citep{vieira2020supply, van2024tabular, mishra2018big}, generated continuously by enterprise resource planning (ERP) systems, warehouse and transportation management systems, and production planning modules \citep{cheng2022linkages, luo2022data}. These systems record every aspect of operations, spanning procurement, inventory, manufacturing, and logistics, in relational schemas whose columns are deeply interdependent. A typical dataset spans purchase orders, shipment records, inventory transactions, and production orders, each encoding rich hierarchical, temporal, and mathematical constraints such as supplier--material mappings, order--delivery orderings, and bill-of-materials dependencies \citep{proselkov2024financial, khayyati2022machine}.

What makes supply chain tabular data particularly challenging for generative modelling is its inherent complexity. Unlike unstructured modalities such as text and images, which follow well-defined spatial or temporal distributions, tabular data feature both categorical and continuous columns with rich inter-column dependencies that lack clear structural patterns \citep{shan2019crowdsourcing, wang2024harmonic, van2024tabular}. Each column may follow its own distinct distribution, and the relationships between columns are often governed by domain-specific logic rather than simple statistical correlations \citep{long2025evaluating, xu2024llms}. This complexity is further amplified in production contexts, where tables are high-dimensional, heterogeneous, and linked by operational rules that vary across firms and sectors \citep{vlachos2025machine, cheng2022linkages}.

The utility of this data in production research is extensive, underpinning tasks ranging from demand forecasting \citep{arora2025forecasting} and predictive maintenance \citep{schneckenreither2021order} to supply chain risk prediction \citep{wyrembek2025causal, zheng2023federated}, payment delay prediction \citep{kong2026hierarchical}, and fraud detection \citep{lokanan2025supply}. Synthetic alternatives have therefore been widely explored as a means to mitigate data scarcity and enable cross-organizational sharing \citep{long2025leveraging, zheng2024blockchain, he2026blockchain}. However, for these synthetic datasets to be reliable in production settings, they must preserve not only statistical properties but also the domain-specific logical relationships embedded in the original schema, a requirement that remains insufficiently addressed in existing tabular generation research.

\subsection{AI-Driven Data Generation in Production Research}
Deep generative methods for tabular data have evolved rapidly along three main directions: GAN-based approaches such as CTGAN \citep{xu2019modeling} and CTAB-GAN+ \citep{zhao2024ctab}; diffusion-based models including TabDDPM \citep{kotelnikov2023tabddpm}, TabSyn \citep{zhang2023mixed}, CoDi \citep{lee2023codi}, and financial-domain variants \citep{sattarov2023findiff}; and LLM-based generators such as GReaT \citep{borisov2022language}, REaLTabFormer \citep{solatorio2023realtabformer}, Tabula \citep{zhao2023tabula}, and HARMONIC \citep{wang2024harmonic}. These methods have demonstrated strong performance on statistical fidelity and privacy preservation, but largely overlook inter-column relationship preservation critical for supply chain applications \citep{long2025evaluating, xu2024llms, wang2024challenges}. Current evaluation frameworks focus on pairwise correlations and multivariate dependencies \citep{margeloiu2024tabebm, alaa2022faithful}, yet fail to assess whether logical constraints such as temporal ordering, mathematical formulas, or hierarchical consistency are satisfied \citep{suh2024timeautodiff, long2025evaluating}.

In parallel, production research has begun to exploit AI-driven reasoning and generation. Large language models have been applied to manufacturing process planning under Industry 5.0 \citep{ni2025llm} and to data augmentation for product design through user-generated content mining \citep{huang2026mining}, while knowledge graphs have been combined with LLMs for fault diagnostic reasoning in industrial settings \citep{ma2025knowledge}. Diffusion models have also entered manufacturing applications through generative optimization for resilient service composition in cloud manufacturing \citep{shahab2026diffusion}. These works highlight LLMs' emerging role as reasoning engines and diffusion models' generative capability in production contexts. However, integrating these promising methods remains challenging because they operate on fundamentally different representations: LLMs express operational knowledge in natural language, whereas diffusion models optimize statistical data distributions without explicit awareness of domain logic. As a result, existing methods have not yet inferred and enforced logical relationships within tabular data for synthetic data generation.

Complementary to synthetic generation, the need for privacy-preserving data sharing in supply chains has been tackled through blockchain-based mechanisms \citep{zheng2024blockchain, he2026blockchain} and federated learning \citep{zheng2023federated}. However, these approaches focus on securely sharing information rather than generating logically consistent synthetic alternatives. Knowledge graphs have meanwhile proven effective for supply chain link prediction and risk management \citep{kosasih2022machine, kosasih2024knowledge, kosasih2025trustworthy, almahri2024enhancing}, demonstrating that structured relational representations can capture complex supply chain logic. However, KG-based reasoning has not yet been applied to the inter-column logic of supply chain tabular data for synthetic generation, which is the gap our work addresses.

\subsection{Towards Logically Consistent Supply Chain Data Generation}

The broader notion of a \textit{world model}, defined as a representation of the rules and logical laws governing a domain, has received growing attention in generative AI as a path beyond pure distributional fidelity \citep{del2025world, kusiak2020convolutional}. While world models have been extensively studied in visual and robotic domains, comparable representations for supply chain tabular data generation remain underdeveloped. Unlike visual domains where constraints are spatially apparent, supply chain logic is encoded implicitly across column relationships that are abstract, high-dimensional, and context-dependent \citep{long2025evaluating, xu2024llms}. A single production table may simultaneously embed hierarchical taxonomies, arithmetic formulas, temporal sequences, and conditional operational rules, none of which are captured by standard distributional metrics \citep{margeloiu2024tabebm, suh2024timeautodiff}.

This challenge is compounded by the fact that inter-column relationships vary substantially across datasets: a retail schema may enforce geographical hierarchies and pricing formulas, whereas a procurement schema may enforce supplier--material mappings and delivery lead-time constraints \citep{luo2022data, proselkov2024financial}. No single hard-coded rule set can generalize across schemas, motivating methods that can automatically discover such structure from metadata and data. Recent work on structure-aware tabular generation such as GOGGLE \citep{liu2023goggle}, which learns relational structure for generative modelling on benchmark data, points in this direction but does not target supply chain constraints or introduce LLM reasoning with knowledge graph representations. To the best of our knowledge, this work represents the first attempt to automatically discover and enforce the inter-column logical structure of supply chain tabular data through LLM-driven knowledge graph reasoning \citep{kosasih2024knowledge, kosasih2025trustworthy}, capturing dataset-specific constraints without manual rule engineering.

\section{TabKG Approach}
\label{sec:TabKG}

\begin{figure*}[t!]
    \centering
    \includegraphics[width=1\textwidth]{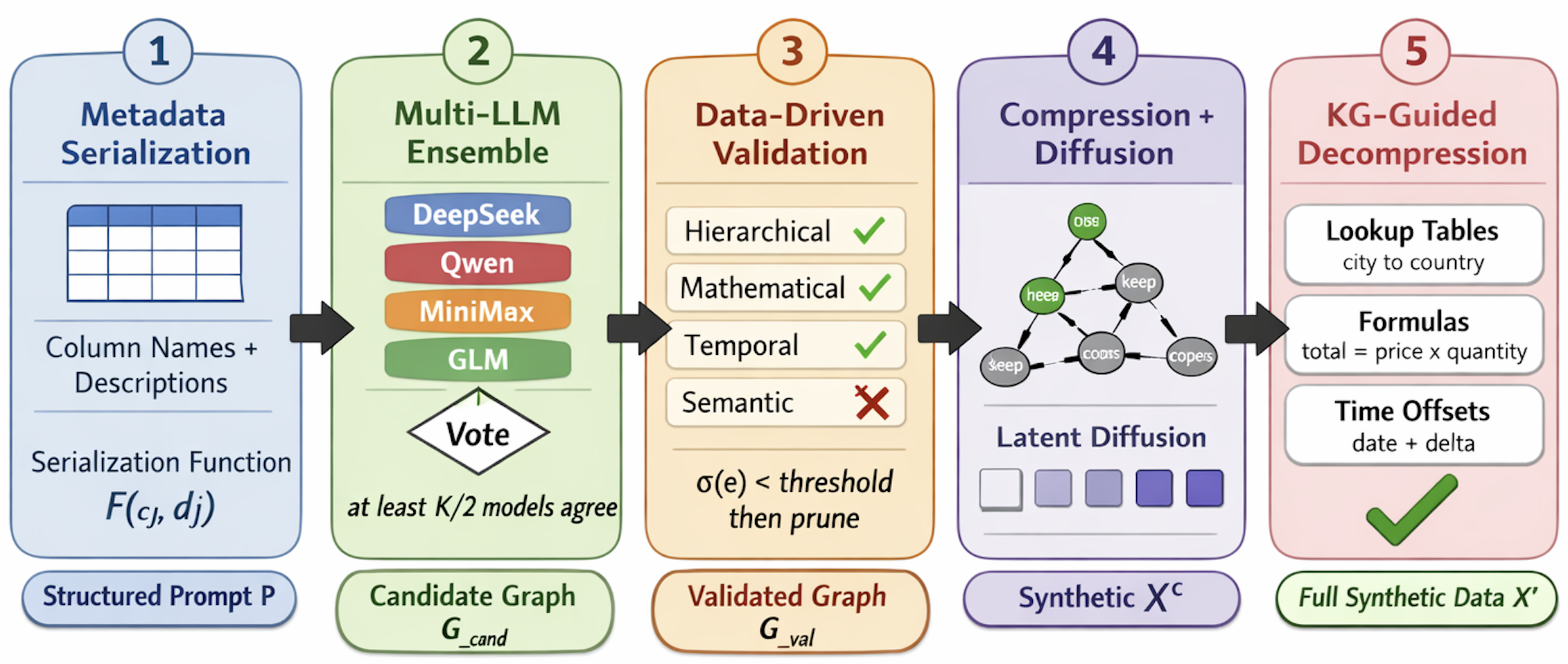}
    \caption{Overview of the TabKG framework. Stage~1 serializes column metadata. Stage~2 constructs a candidate knowledge graph via a multi-LLM ensemble with majority voting. Stage~3 validates each edge against real data and prunes hallucinated relationships. Stage~4 uses the validated graph for compression and latent diffusion-based generation. Stage~5 reconstructs the full synthetic dataset via knowledge graph-guided decompression.}
    \label{fig:TabKG}
\end{figure*}

TabKG explicitly constructs a Column Relationship Knowledge Graph (CR-KG) that captures inter-column dependencies, uses this graph to compress the dataset to its minimal independent representation, generates synthetic data in this compressed space via a latent diffusion model, and reconstructs the full dataset by reapplying the logical constraints. Since the reconstruction functions are deterministic and derived from the validated knowledge graph, all \emph{discovered} logical relationships are satisfied \textit{by construction}; relationships that the CR-KG fails to discover are not enforced as hard constraints and are instead handled implicitly by the latent diffusion model. The framework operates through five stages, as illustrated in \autoref{fig:TabKG}: (1) metadata serialization, (2) multi-LLM ensemble graph construction with majority voting, (3) data-driven validation and pruning, (4) graph-guided compression and diffusion-based generation, and (5) knowledge graph-guided decompression.

\paragraph{Stage 1: Metadata Serialization.}
Let $\mathbf{X} \in \mathbb{R}^{m \times n}$ denote a tabular dataset with $m$ rows and $n$ columns indexed by $j \in \{1, \dots, n\}$, and let $\mathbf{x}_j \in \mathbb{R}^N$ denote the values of the $j$-th column. Each column $j$ has a name $c_j$ and a description $d_j$, collected as $C = [c_1, \dots, c_n]$ and $D = [d_1, \dots, d_n]$. For each $j$, the pair $(c_j, d_j)$ is serialized into natural language via $F:(c_j, d_j) \mapsto \texttt{``}c_j \texttt{:} d_j\texttt{''}$, and the resulting strings are combined into a structured prompt $\mathcal{P}$ that instructs the LLM to output inter-column relationships as a typed knowledge graph.

\begin{figure*}[t]
    \centering
    \includegraphics[width=0.95\textwidth]{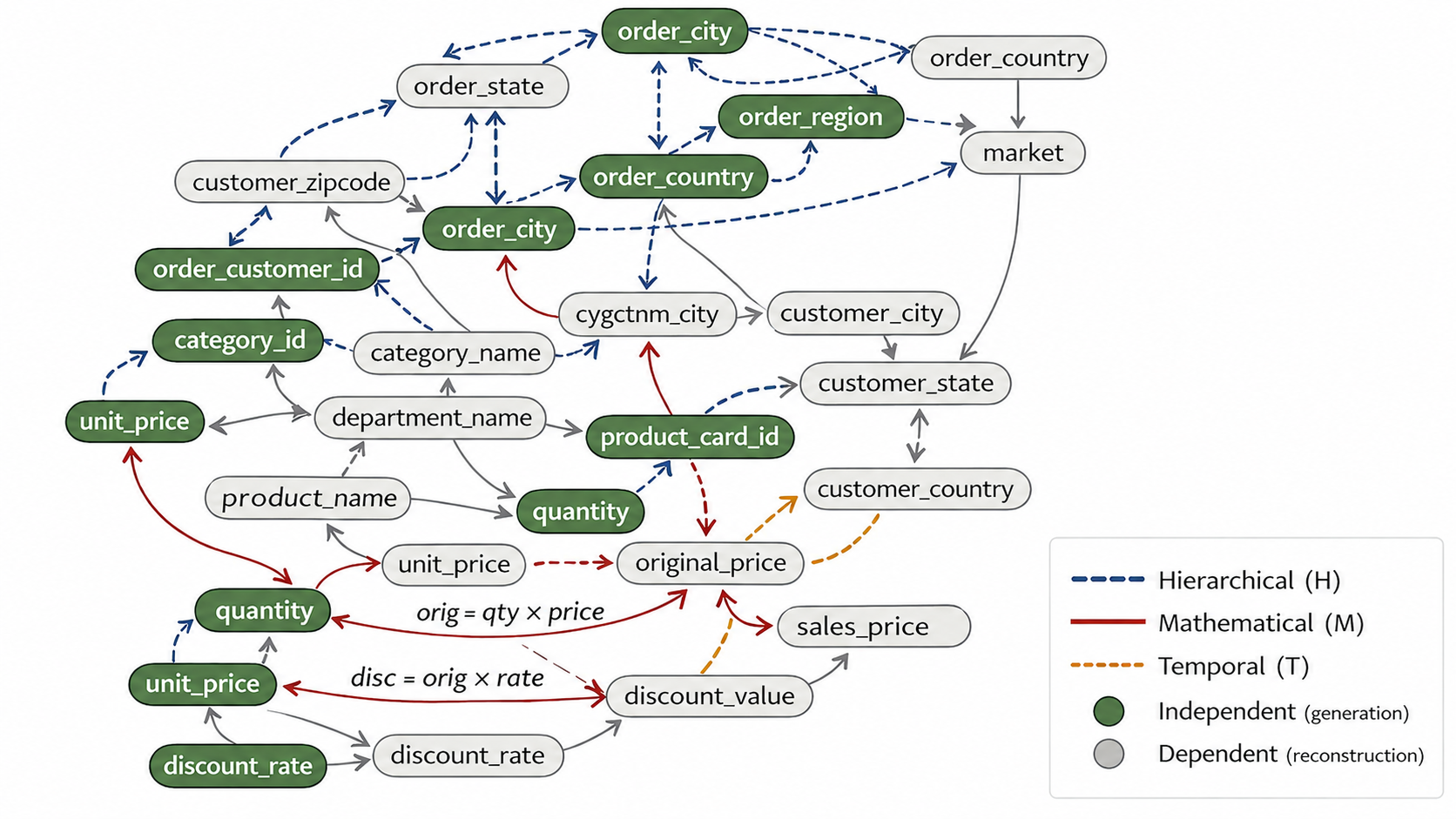}
    \caption{Example Column Relationship Knowledge Graph (CR-KG) for the Retail dataset. Bold-bordered nodes represent independent columns retained for generation; faded nodes are compressed away and reconstructed during decompression.}
    \label{fig:crkg_example}
\end{figure*}

\paragraph{Stage 2: Multi-LLM Ensemble Graph Construction.}\label{sec:ensemble}
We define the Column Relationship Knowledge Graph as a directed typed graph $G = (V, E)$, where each node $v_j \in V$ corresponds to the $j$-th column ($|V| = n$), and each edge $e = (v_s, v_t, \tau, \rho, r) \in E$ encodes a relationship from a source column $v_s$ to a target column $v_t$ with type $\tau \in \{H, M, T, S\}$ (hierarchical, mathematical, temporal, or semantic), confidence $\rho \in [0,1]$, and rule $r$. An example CR-KG for the Retail dataset is shown in \autoref{fig:crkg_example}.

Rather than relying on a single LLM, we employ $K$ models $\{\mathcal{M}_1, \dots, \mathcal{M}_K\}$ with majority voting. Each model $\mathcal{M}_k$ independently produces a candidate graph $G_k = (V, E_k)$, and an edge is retained only if proposed by at least $\lceil K/2 \rceil$ of the models:
\begin{equation}
    E_{\text{cand}} = \bigl\{ e : \textstyle\sum_{k=1}^{K} \mathbb{1}[ e \in E_k ] \geq \lceil K/2 \rceil \bigr\},
\end{equation}
where $\mathbb{1}[\cdot]$ is the indicator function. The resulting candidate graph is $G_{\text{cand}} = (V, E_{\text{cand}})$.

\paragraph{Stage 3: Data-Driven Edge Validation.}\label{sec:validation}
The candidate graph $G_{\text{cand}}$ may contain hallucinated edges. We validate each edge $e = (v_s, v_t, \tau, \rho, r)$ against the actual data $\mathbf{X}$ using type-specific validators that compute a score $\sigma(e) \in [0,1]$: hierarchical edges are validated by checking functional dependency (whether the source column values $\mathbf{x}_s$ uniquely determine the target column values $\mathbf{x}_t$), mathematical edges by evaluating formula accuracy on real rows, temporal edges by checking ordering violations, and semantic edges by evaluating conditional rule satisfaction rates (e.g., checking whether a product description consistently maps to its assigned category). Edges with $\sigma(e) < \theta$ (default $\theta = 0.90$) are pruned, yielding the validated graph $G_{\text{val}} = (V, E_{\text{val}})$. The hallucination rate $(|E_{\text{cand}}| - |E_{\text{val}}|)/|E_{\text{cand}}|$ is reported.

\paragraph{Stage 4: Graph-Guided Compression and Generation.}\label{sec:compression}
The validated graph $G_{\text{val}}$ is converted into a directed acyclic graph (DAG) and used to identify the minimal independent column set $V_{\text{keep}} = \{v \in V : \text{in-degree}(v) = 0\}$, with the dependent set $V_{\text{compress}} = V \setminus V_{\text{keep}}$. Each dependent column $v \in V_{\text{compress}}$ is associated with a reconstruction function $f_v$ whose form is determined by the type of the incoming edge: lookup tables for hierarchical and semantic edges, closed-form formulas for mathematical edges, and additive time offsets for temporal edges. We collectively denote the set of reconstruction functions as $\mathcal{F} = \{f_v : v \in V_{\text{compress}}\}$.

We define the row-wise compression operator $\mathcal{C}: \mathbb{R}^n \to \mathbb{R}^{|V_{\text{keep}}|}$, which projects a full row $\mathbf{x}$ onto the columns in $V_{\text{keep}}$, and the row-wise decompression operator $\mathcal{D}(\cdot;\mathcal{F},G_{\text{val}}): \mathbb{R}^{|V_{\text{keep}}|} \to \mathbb{R}^n$, which restores all dependent columns by applying $\mathcal{F}$ in topological order over $G_{\text{val}}$. For deterministic relationships ($\sigma(e) = 1.0$), this compression--decompression is provably lossless: $\mathcal{D}(\mathcal{C}(\mathbf{x});\mathcal{F},G_{\text{val}}) = \mathbf{x}$ for every row $\mathbf{x}$ of $\mathbf{X}$. Applying $\mathcal{C}$ row-wise to $\mathbf{X}$ yields the compressed dataset $\mathbf{X}_{\text{c}} \in \mathbb{R}^{N \times |V_{\text{keep}}|}$, which is then used to train a latent score-based diffusion model \citep{zhang2023mixed, karras2022elucidating} that generates synthetic compressed data $\mathbf{X}'_{\text{c}}$.

\paragraph{Stage 5: Knowledge Graph-Guided Decompression.}
The compressed synthetic data $\mathbf{X}'_{\text{c}}$ is decompressed into the full synthetic dataset $\mathbf{X}'$ by applying $\mathcal{D}(\cdot;\mathcal{F},G_{\text{val}})$ row-wise; equivalently, $\mathbf{X}' = \mathcal{D}(\mathbf{X}'_{\text{c}};\mathcal{F},G_{\text{val}})$. Each reconstruction function $f_d \in \mathcal{F}$ is applied in topological order over the DAG: hierarchical and semantic columns via lookup tables, mathematical columns via validated formulas, and temporal columns by adding generated offsets to base timestamps. Because all reconstruction functions are deterministic and derived from validated edges, logical consistency \emph{on the validated edges} is guaranteed by construction; relationships that are absent from $G_{\text{val}}$ are not enforced and are left to the diffusion model.

The complete TabKG pipeline is summarised in \autoref{alg:tabkg}.

\begin{algorithm}[t]
\caption{TabKG: Knowledge Graph-Guided Synthetic Data Generation}
\label{alg:tabkg}
\DontPrintSemicolon
\KwIn{Dataset $\mathbf{X}$, column metadata $(C, D)$, LLMs $\{\mathcal{M}_1, \dots, \mathcal{M}_K\}$, threshold $\theta$}
\KwOut{Synthetic dataset $\mathbf{X}'$ with preserved logical relationships}
\BlankLine
\tcp{Stage 1: Metadata Serialization}
$\mathcal{P} \leftarrow \text{SerializePrompt}(C, D)$\;
\BlankLine
\tcp{Stage 2: Multi-LLM Ensemble}
\For{$k = 1$ \KwTo $K$}{
    $G_k \leftarrow \mathcal{M}_k(\mathcal{P})$ \tcp*{Each LLM proposes a candidate graph}
}
$G_{\text{cand}} \leftarrow \text{MajorityVote}(\{G_1, \dots, G_K\}, \lceil K/2 \rceil)$\;
\BlankLine
\tcp{Stage 3: Data-Driven Validation}
\For{each edge $e \in G_{\text{cand}}$}{
    $\sigma(e) \leftarrow \text{Validate}(e, \mathbf{X})$ \tcp*{Type-specific validation}
}
$G_{\text{val}} \leftarrow \{e \in G_{\text{cand}} : \sigma(e) \geq \theta\}$\;
\BlankLine
\tcp{Stage 4: Graph-Guided Compression + Diffusion}
$(V_{\text{keep}}, V_{\text{compress}}, \mathcal{F}) \leftarrow \text{DAGCompress}(G_{\text{val}})$\;
$\mathbf{X}_{\text{c}} \leftarrow \mathcal{C}(\mathbf{X}; V_{\text{keep}})$ \tcp*{Keep only independent columns}
$\mathbf{X}'_{\text{c}} \leftarrow \text{LatentDiffusion}(\mathbf{X}_{\text{c}})$ \tcp*{Generate via score-based diffusion}
\BlankLine
\tcp{Stage 5: KG-Guided Decompression}
$\mathbf{X}' \leftarrow \mathcal{D}(\mathbf{X}'_{\text{c}}; \mathcal{F}, G_{\text{val}})$ \tcp*{Reconstruct dependent columns}
\Return{$\mathbf{X}'$}
\end{algorithm}

\section{Experimental Settings}
\label{sec:experimental_settings}

\subsection{Datasets}
\label{sec:dataset}

\begin{table}[th]
\centering
\caption{Statistics of the two evaluation datasets for synthetic data generation tasks.}
\label{tab:exp-dataset}
\renewcommand{\arraystretch}{1.15}
\resizebox{0.7\textwidth}{!}{%
\begin{tabular}{rccc|cc}
\toprule
Dataset &
  \begin{tabular}[c]{@{}c@{}}\# of \\ Rows\end{tabular} &
  \begin{tabular}[c]{@{}c@{}}\# of \\ Num. Feat.\end{tabular} &
  \begin{tabular}[c]{@{}c@{}}\# of \\ Cat. Feat.\end{tabular} &
  \begin{tabular}[c]{@{}c@{}}Size of \\ Training Set\end{tabular} &
  \begin{tabular}[c]{@{}c@{}}Size of \\ Test Set\end{tabular} \\
\midrule
Retail &
  172,765 &
  26 &
  15 &
  155,488 &
  17,277 \\
Purchasing &
  29,590 &
  14 &
  7 &
  26,631 &
  2,959 \\
\bottomrule
\end{tabular}%
}
\end{table}

We evaluate TabKG on two real-world industrial supply chain datasets: a public Retail dataset~\citep{dataco} and a proprietary Purchasing dataset. These datasets are selected because they contain complex inter-column relationships, including dependencies among lead times, order cycles, product and geographical hierarchies, supplier information, and financial attributes. Such relationships make them suitable for evaluating whether synthetic tabular data preserves not only marginal and joint statistical distributions, but also the operational logic of the underlying supply chain system.

As shown in \autoref{tab:exp-dataset}, the Retail dataset contains 172,765 records and 41 features, including 26 numerical and 15 categorical features. It covers geographical hierarchies, product taxonomies, and financial calculations, and is used for the downstream task of \emph{late-delivery risk prediction}, a standard service-level-agreement monitoring task in logistics planning. The Purchasing dataset contains 29,590 records and 21 features, including 14 numerical and 7 categorical features. It captures procurement cycles, supplier relationships, delivery timelines, and tax-inclusive transactional values, and is used for the downstream task of \emph{procurement-status classification}, an order-to-cash monitoring task. The full set of inter-column relationships discovered by TabKG, spanning hierarchical, mathematical, temporal, and semantic types, is reported and analysed in \autoref{tab:relationships} of \autoref{sec:logic_results}.

\subsection{Baselines}
\label{sec:baselines}

We compare TabKG against four representative state-of-the-art tabular generative baselines spanning multiple modelling paradigms. CTGAN~\citep{xu2019modeling} is included as a widely adopted GAN-based framework for tabular data synthesis. TabDDPM~\citep{kotelnikov2023tabddpm} is included as a foundational diffusion model for tabular data. TabSyn~\citep{zhang2023mixed} is included as a recent diffusion-based method for mixed-type tabular generation. GReaT~\citep{borisov2022language} is included as an LLM-based method that reformulates tabular rows as textual sequences to leverage pre-trained language representations. Together, these baselines provide a diverse comparison across adversarial, diffusion-based, and language-model-based approaches to tabular data generation.

For the operational logic reasoning task, we also compare TabKG against a standard prompt-only baseline. This baseline uses separate prompts to infer relationship types directly from table metadata, without constructing a column-reasoning knowledge graph or performing graph-based validation. This comparison isolates the contribution of TabKG's structured CR-KG representation and validation process.

\subsection{Experimental Setup}
\label{sec:experimental_protocol}

We evaluate TabKG through two complementary experimental tasks: operational logic reasoning and synthetic tabular data generation.

For the operational logic reasoning task, each LLM is provided with table metadata, including column names and column descriptions, and is asked to infer inter-column relationships through the CR-KG structured prompt. We benchmark four diverse LLMs from different model families: DeepSeek~\citep{deepseekai2025}, Qwen-3.5~\citep{qwen2024qwen25}, MiniMax-M2.5~\citep{minimax2025minimax}, and GLM-5~\citep{glm2024chatglm}. We evaluate two TabKG ensemble configurations. The first is a \textit{cross-model ensemble}, which applies majority voting across the three strongest models, namely DeepSeek, Qwen-3.5, and MiniMax. The second is a \textit{same-model ensemble}, which queries DeepSeek five times with temperatures $\tau \in \{0.1, 0.2, 0.3, 0.4, 0.5\}$ and applies majority voting to assess robustness to decoding stochasticity. The inferred relationships are evaluated against manually annotated ground-truth relationships.

\begin{figure*}[t]
    \centering
    \includegraphics[width=\textwidth]{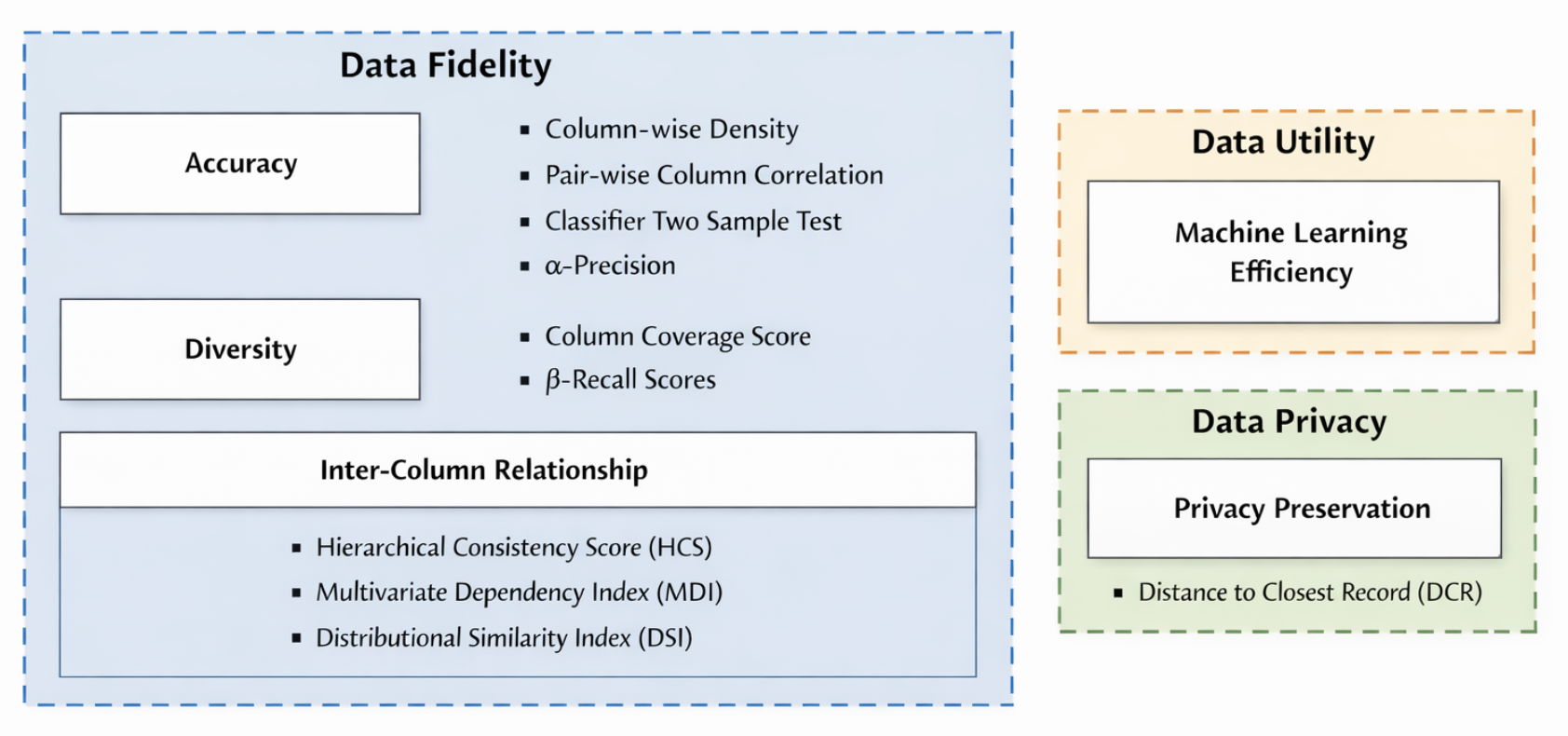}
    \caption{Evaluation framework for synthetic tabular data in supply chain settings.}
    \label{fig:evaluation_framework}
\end{figure*}

For the synthetic data generation task, each method generates ten independent synthetic datasets for each real dataset, and we report the mean and standard deviation across runs. As illustrated in \autoref{fig:evaluation_framework}, the evaluation follows a multi-dimensional framework covering data fidelity, data utility, data privacy, and inter-column logical consistency. This design allows us to assess whether TabKG improves logical validity without sacrificing the standard desiderata of synthetic tabular data generation.

\subsection{Evaluation Metrics}
\label{sec:evaluation_metrics}

We use separate evaluation metrics for the two experimental tasks: operational logic reasoning and synthetic tabular data generation.

\subsubsection{Operational Logic Reasoning Metrics}
\label{sec:logic_discovery_metrics}

For the column-reasoning task, we evaluate inferred inter-column relationships using precision, recall, and F1 score against manually annotated ground-truth relationships. Precision measures the fraction of inferred relationships that are correct, recall measures the fraction of ground-truth relationships recovered by the model, and F1 summarizes the balance between precision and recall. These metrics are reported for each LLM, for the prompt-only baseline, and for the two TabKG ensemble configurations. This evaluation measures whether the model can correctly identify operational rules such as hierarchical mappings, mathematical dependencies, temporal constraints, and semantic relationships among columns.

\subsubsection{Synthetic Data Generation Metrics}
\label{sec:generation_metrics}

For the synthetic data generation task, we evaluate generated datasets along four dimensions: fidelity, logical consistency, utility, and privacy. Fidelity measures whether the generated records match the statistical distribution of the real data. Logical consistency measures whether generated records preserve operational rules across columns. Utility measures whether synthetic data can support downstream predictive modelling. Privacy measures whether generated records avoid direct memorization of real records.

\paragraph{Data fidelity.}
Following the evaluation framework in~\citep{long2025evaluating}, we assess data fidelity from both statistical and logical perspectives. Statistical fidelity is measured using column-wise density estimation, pairwise correlation similarity, and $\alpha$-Precision~\citep{alaa2022faithful}, which evaluate whether synthetic records preserve univariate feature distributions and multivariate statistical structure. Logical fidelity is measured using three inter-column consistency metrics from prior work~\citep{long2025evaluating}: the Hierarchical Consistency Score (HCS), which evaluates whether hierarchical mappings such as city-to-country or product-to-category relationships are preserved; the Multivariate Dependency Index (MDI), which measures the satisfaction rate of mathematical and temporal constraints; and the Distributional Similarity Index (DSI), which captures implicit distributional dependencies across related columns. Together, these metrics test whether synthetic supply chain data preserves not only statistical realism, but also the operational rules that govern the real system.

\paragraph{Data utility.}
Data utility measures whether synthetic data can support downstream predictive modelling. We use the Train-on-Synthetic, Test-on-Real (TSTR) protocol. For each generated dataset, an XGBoost classifier~\citep{chen2016xgboost} is trained on synthetic data and evaluated on the held-out real test set. We report AUC and F1 score for the two downstream tasks: late-delivery risk prediction on the Retail dataset and procurement-status classification on the Purchasing dataset.

\paragraph{Data privacy.}
Data privacy measures whether synthetic records reveal or closely replicate real records. We use Distance to Closest Record and Classifier Two-Sample Test~\citep{liu2024scaling}. Distance to Closest Record evaluates the proximity between generated samples and real records, while Classifier Two-Sample Test measures whether a classifier can distinguish real from synthetic data. Together, these metrics assess whether the synthetic data avoids direct memorization while remaining distributionally realistic.

\subsection{Hyperparameter Optimization}
\label{sec:hyperparameters}

To ensure fair downstream utility evaluation, we conduct a grid search for the XGBoost classifier on the validation split. The search space includes the number of trees $\{100, 200, 300\}$, learning rates $\{0.1, 0.01, 0.001\}$, maximum tree depths $\{5, 10\}$, minimum child weights $\{1, 10\}$, and gamma values $\{0.0, 0.5, 1.0\}$. The best-performing configuration is then used for the TSTR evaluation.

For the baseline generative models, including CTGAN, TabDDPM, TabSyn, and GReaT, we follow the hyperparameter configurations recommended in their original publications. This ensures a standardized comparison and avoids tuning baselines specifically against TabKG. All experiments are conducted using the optimal downstream classifier configuration identified during validation.

\section{Experimental Results}

\subsection{Logic Reasoning Results}\label{sec:logic_results}

\begin{table*}[t]
\centering
\caption{Column reasoning performance on the two supply chain datasets.
\textit{Baseline}: a standard single-prompt approach applied to each LLM independently without knowledge graph construction.
\textit{TabKG (ours)}: the full CR-KG pipeline with majority voting, in which cross-model uses DeepSeek, Qwen-3.5, and MiniMax, while same-model queries DeepSeek five times at temperatures 0.1--0.5.
The best and second-best results are highlighted in {\color{blue}\bf blue} and {\bf bold}, respectively.}
\label{tab:reasoning}
\renewcommand{\arraystretch}{1.15}
\small
\resizebox{\textwidth}{!}{%
\begin{tabular}{llcccccc}
\toprule
& & \multicolumn{4}{c}{\textbf{Baseline (Single Prompt, No KG)}} & \multicolumn{2}{c}{\textbf{TabKG (Ours)}} \\
\cmidrule(lr){3-6} \cmidrule(lr){7-8}
\textbf{Dataset} & \textbf{Metric} & \textbf{DeepSeek} & \textbf{Qwen-3.5} & \textbf{MiniMax} & \textbf{GLM-5} & \textbf{Cross-Model} & \textbf{Same-Model} \\
\midrule
\multirow{3}{*}{\textbf{Retail}}
& Precision & 0.600 & 0.556 & 0.472 & 0.500 & \textbf{0.941} & \textcolor{blue}{\textbf{0.971}} \\
& Recall    & 0.500 & 0.179 & 0.393 & 0.321 & 0.444 & \textcolor{blue}{\textbf{0.833}} \\
& F1        & 0.545 & 0.271 & 0.429 & 0.393 & \textbf{0.604} & \textcolor{blue}{\textbf{0.897}} \\
\midrule
\multirow{3}{*}{\textbf{Purchasing}}
& Precision & 0.385 & 0.600 & 0.462 & 0.500 & \textcolor{blue}{\textbf{0.976}} & \textbf{0.962} \\
& Recall    & 0.357 & 0.462 & 0.462 & 0.385 & \textcolor{blue}{\textbf{0.971}} & \textbf{0.929} \\
& F1        & 0.370 & 0.522 & 0.462 & 0.435 & \textcolor{blue}{\textbf{0.973}} & \textbf{0.945} \\ \bottomrule
\end{tabular}%
}
\end{table*}

We evaluate and compare the column-logic understanding performance of TabKG against baseline methods that use a single text prompt without knowledge graph construction, by measuring the inter-column relationships each approach infers against manually annotated ground-truth relationships for both supply chain datasets. The ground truth comprises 36 relationships for the Retail dataset and 14 for the Purchasing dataset, covering hierarchical, mathematical, temporal, and semantic edge types.
\autoref{tab:reasoning} presents the reasoning performance across four LLMs (DeepSeek, Qwen-3.5, MiniMax, and GLM-5) representing different model families and providers. We treat \autoref{tab:reasoning} as an \textit{ablation of the CR-KG construction pipeline}: the \textbf{Baseline (Single Prompt, No KG)} columns isolate the contribution of a single LLM with standard prompting and no structured knowledge graph, while the \textbf{TabKG (Ours)} columns progressively add (i) the CR-KG structured prompt and (ii) majority-voting ensembling, either across different LLMs (\textit{cross-model}) or across temperatures within a single LLM (\textit{same-model}). This allows us to attribute the performance gain to the CR-KG prompt and to the ensemble consensus mechanism separately.

As shown, the no-KG baseline achieves F1 scores of only 0.27--0.55 on Retail and 0.37--0.52 on Purchasing, with no single model consistently performing well. Both TabKG ensemble configurations substantially outperform this ablation. The \textbf{cross-model ensemble} (majority voting across DeepSeek, Qwen-3.5, and MiniMax) achieves the best F1 of 0.973 on Purchasing, recovering almost all ground-truth relationships with very few false positives. The \textbf{same-model ensemble} (DeepSeek queried five times at temperatures 0.1--0.5) achieves the best performance on Retail, with precision of 0.971 and F1 of 0.897. The remaining missed edges correspond to indirect relationships difficult to infer from column descriptions alone. The gap between single-prompt baseline and either ensemble (F1 improvement of +0.35 to +0.51 absolute) demonstrates that both the CR-KG structured prompt and the consensus voting are necessary components.

Importantly, both TabKG variants achieve precision $\geq$ 0.92 across all experiments, demonstrating that the CR-KG structured prompt combined with ensemble voting effectively mitigates LLM hallucination. Moreover, an F1 below 1.0 does not translate into catastrophic generation failure: hallucinated edges are filtered by the data-driven validation stage (\autoref{sec:validation}) and only survive if they actually hold on the data, in which case enforcing them is harmless; missed relationships are not enforced as hard constraints and fall back to the regime of the underlying latent diffusion baseline (e.g., TabSyn) rather than producing constraint-violating outputs. \autoref{tab:logic}--\autoref{tab:utility} confirm this: even when CR-KG construction is imperfect (Retail F1 = 0.90), the resulting synthetic data still substantially outperforms TabSyn on logical consistency.

\subsection{Data Generation Results}\label{sec:results}

This section presents the evaluation results for synthetic data produced by TabKG and baseline methods. TabKG uses the validated CR-KG obtained from the same-model ensemble (DeepSeek queried five times) to compress the dataset before training the latent diffusion model. Each evaluation is conducted ten times; results are reported as mean $\pm$ standard deviation.

Importantly, the inclusion of \textbf{TabSyn} alongside TabKG constitutes an \textit{ablation of the CR-KG-guided compression stage}: TabKG and TabSyn share the same latent score-based diffusion architecture~\citep{zhang2023mixed, karras2022elucidating}, with the only difference being that TabSyn is trained on the \emph{full} column set whereas TabKG is trained on the \emph{compressed independent column set} identified by $G_{\text{val}}$ (Section~\ref{sec:compression}). Any gap between TabSyn and TabKG in the following tables therefore isolates the contribution of the CR-KG-guided compression and decompression to logical consistency, utility, and privacy.

\subsubsection{Data Fidelity: Accuracy and Diversity}
\label{sec:acc_div}

\renewcommand{\arraystretch}{1.15}
\begin{table}[h]
\centering
\caption{
Evaluation results of the five methods in terms of accuracy (density estimation, Pearson correlation, and $\alpha$-Precision) and diversity (average coverage score and $\beta$-Recall).
Higher values of these metrics indicate better performance.
The best and second-best results are highlighted in {\color{blue}\bf blue} and {\bf black boldfaced}, respectively.
}
\label{tab:accuracy}
\small
\resizebox{\textwidth}{!}{%
\begin{tabular}{ccccccc}
\toprule[1pt]
\multirow{2}{*}{\textbf{Datasets}} &
\multirow{2}{*}{\textbf{Metrics}} &
  \multicolumn{3}{c}{\textbf{Latent Space-based}} &
  \textbf{LLM-based} & \textbf{Ours}\\ \cmidrule(lr){3-5} \cmidrule(lr){6-6} \cmidrule(lr){7-7}
 &  &
  \textbf{CTGAN} &
  \textbf{TabDDPM} &
  \textbf{TabSyn} &
  \textbf{GReaT} &
  \textbf{TabKG} \\ \hline

\multirow{5}{*}{\textbf{Retail}} &
Density Estimation &
  90.38±0.23 & 33.11±0.12 & \textbf{96.38±0.24} & 89.58±0.32 & \textcolor{blue}{\textbf{96.46±0.23}} \\
&
Pairwise Correlation &
  74.41±0.24 & 36.78±0.31 & \textcolor{blue}{\textbf{94.81±0.44}} & 71.00±1.36 & \textbf{92.52±0.42} \\
&
$\alpha$-Precision &
  88.90±0.24 & 0.00±0.00 & \textbf{98.48±0.21} & 82.18±0.19 & \textcolor{blue}{\textbf{99.15±0.17}} \\
\cmidrule(lr){2-7}
&
Average Coverage Score &
  82.27±0.34 & 76.23±0.23 & \textcolor{blue}{\textbf{99.52±0.31}} & 92.34±0.45 & \textbf{99.18±0.24} \\
&
$\beta$-Recall &
  1.71±0.21 & 0.00±0.00 & 22.62±0.32 & \textbf{24.05±0.21} & \textcolor{blue}{\textbf{27.85±0.23}} \\
\midrule\midrule
\multirow{5}{*}{\textbf{Purchasing}} &
Density Estimation &
  86.36±0.34 & 48.01±0.44 & \textcolor{blue}{\textbf{98.61±0.35}} & 92.72±0.38 & \textbf{98.14±0.47} \\
&
Pairwise Correlation &
  82.25±0.33 & 51.00±0.34 & \textcolor{blue}{\textbf{98.63±0.20}} & 74.18±3.22 & \textbf{92.34±1.12} \\
&
$\alpha$-Precision &
  73.67±0.22 & 4.16±0.31 & \textbf{96.73±0.37} & 88.00±0.36 & \textcolor{blue}{\textbf{97.29±0.25}} \\
\cmidrule(lr){2-7}
&
Average Coverage Score &
  75.87±0.13 & \textcolor{blue}{\textbf{97.49±0.80}} & 91.85±0.45 & 88.72±2.97 & \textbf{93.92±0.23} \\
&
$\beta$-Recall &
  4.23±0.38 & 0.03±0.04 & 15.30±0.37 & \textcolor{blue}{\textbf{36.16±0.32}} & \textbf{17.32±0.21} \\
\bottomrule
\end{tabular}%
}
\end{table}

\begin{figure*}[th]
    \centering
    \begin{subfigure}{0.32\textwidth}
        \centering
        \includegraphics[width=\textwidth]{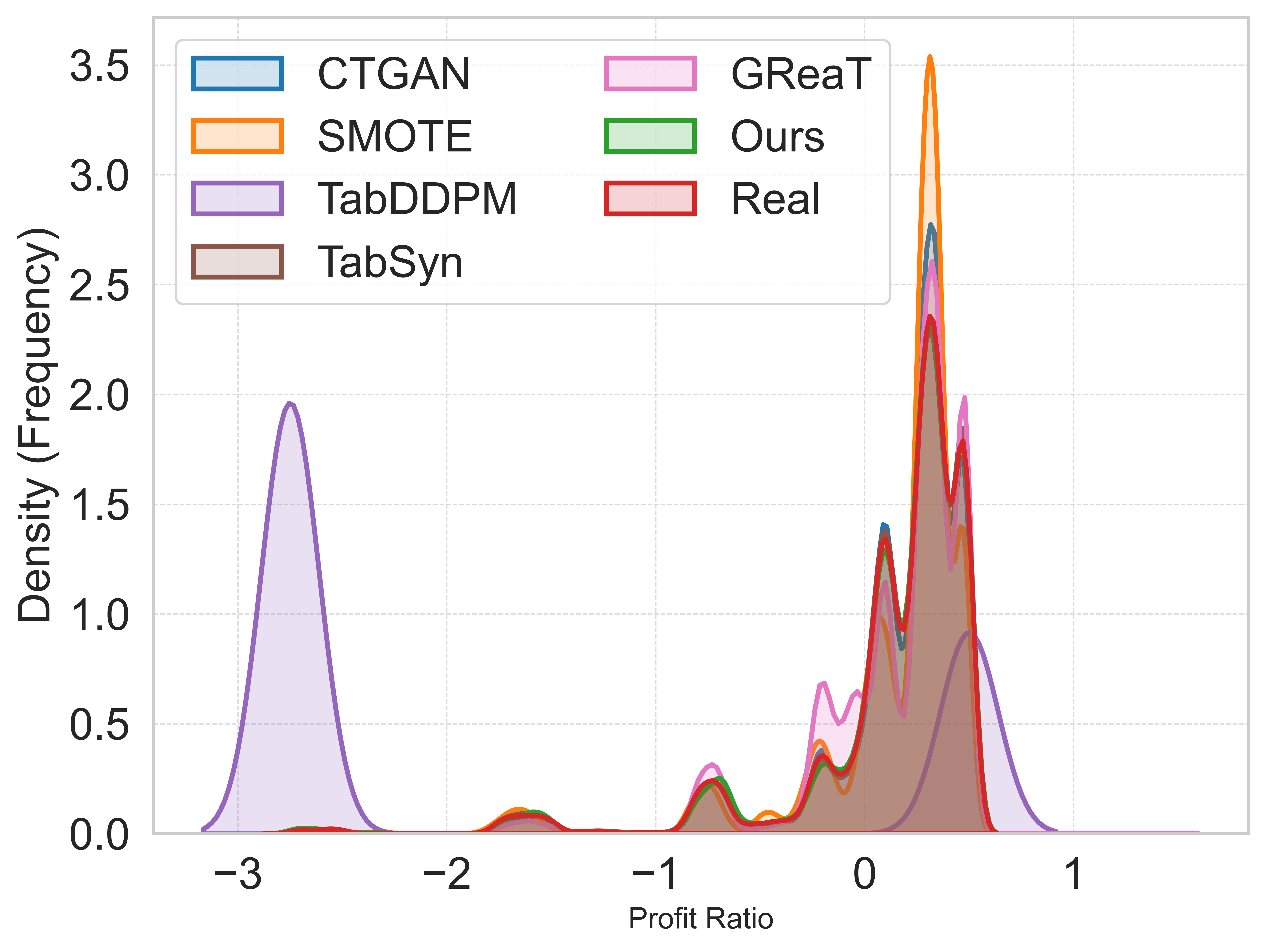}
        \caption{Item Profit Ratio}
    \end{subfigure}
    \begin{subfigure}{0.33\textwidth}
        \centering
        \includegraphics[width=\textwidth]{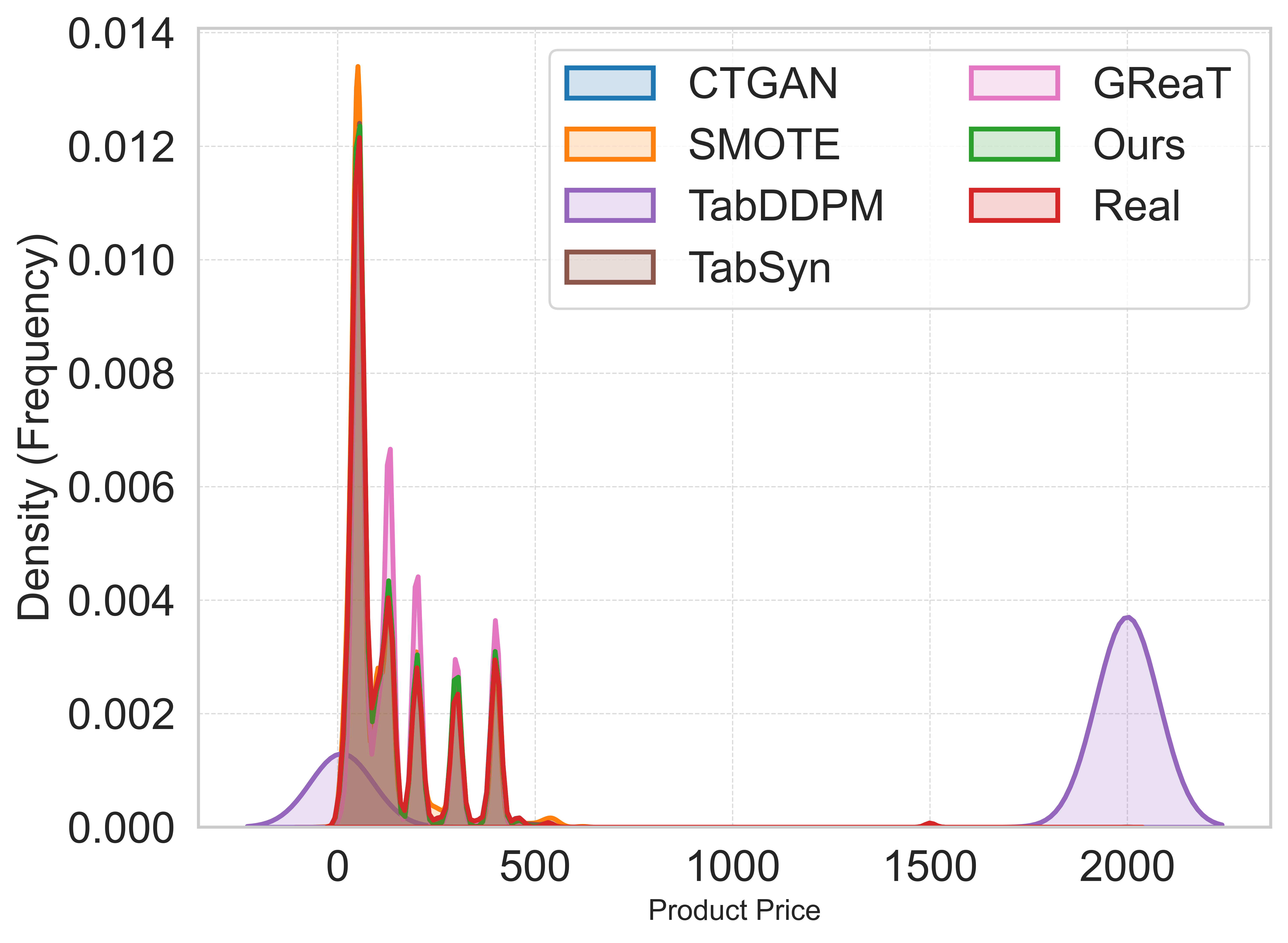}
        \caption{Product Price}
    \end{subfigure}
    \begin{subfigure}{0.325\textwidth}
        \centering
        \includegraphics[width=\textwidth]{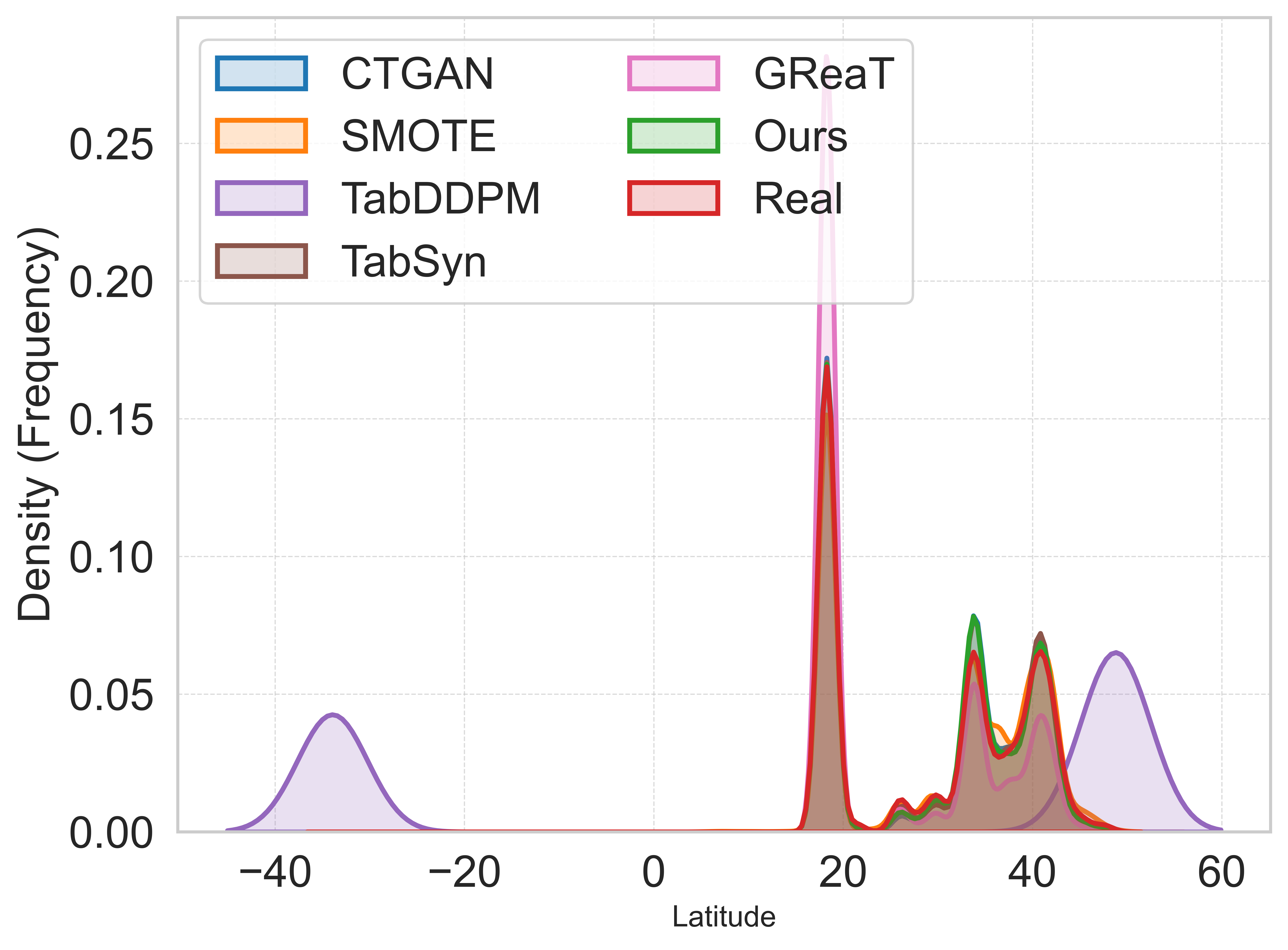}
        \caption{Latitude}
    \end{subfigure}
    \caption{Density plots for the three {\bf continuous} columns ({\tt item profit ratio}, {\tt product price}, and {\tt latitude}), comparing the distribution of real data and their synthetic counterparts generated by CTGAN, TabDDPM, GReaT, TabSyn, and TabKG.
    Curves that more closely align with the real data indicate better performance.
    Both TabKG and TabSyn exhibit distributions that closely match the real data, outperforming other methods.}
    \label{fig:accuracy1}
\end{figure*}

\begin{figure*}[ht!]
    \centering
    \begin{subfigure}[t]{0.32\textwidth}
        \centering
        \includegraphics[width=\textwidth, height=5cm]{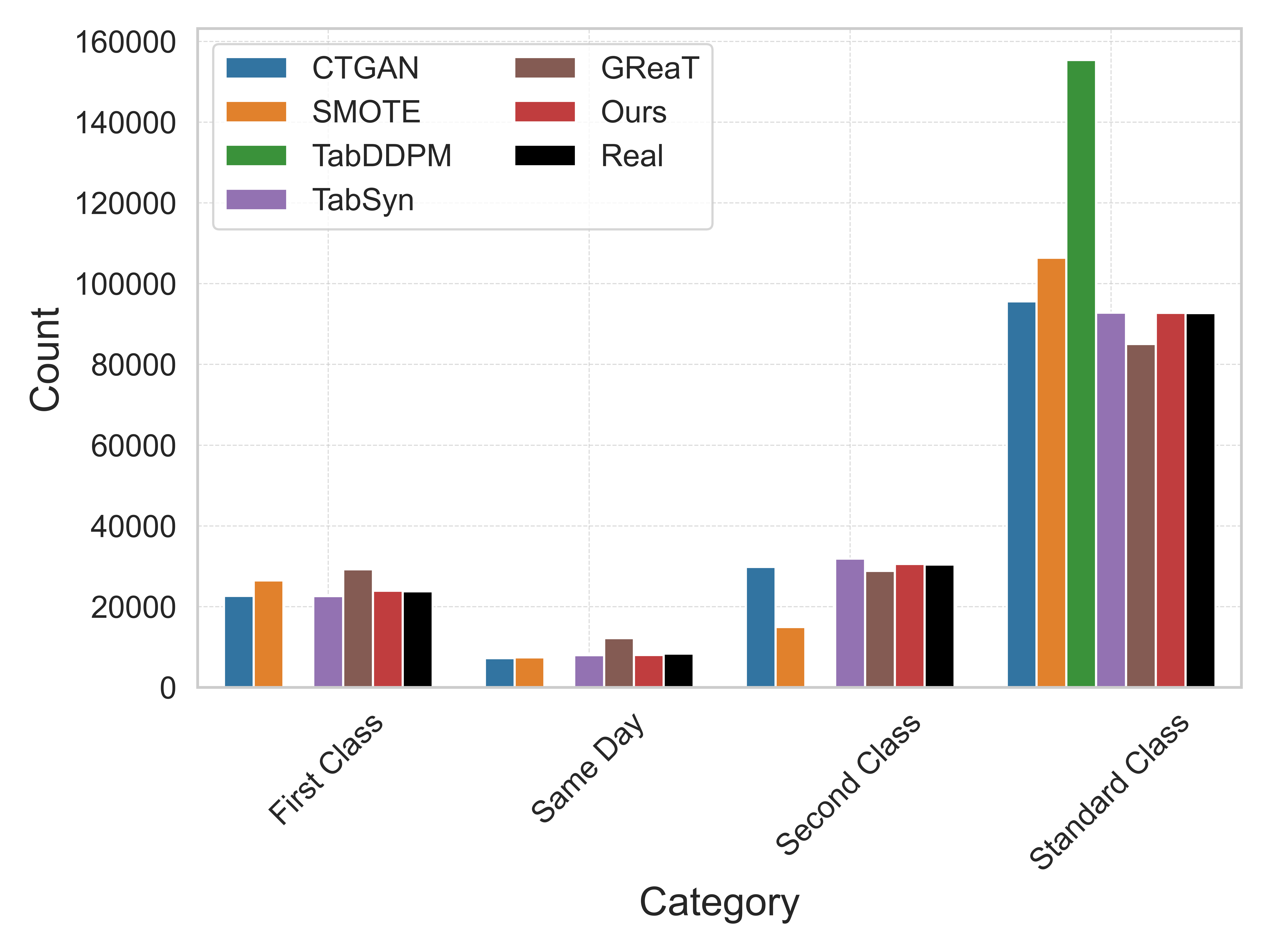}
        \caption{Shipping Mode}
    \end{subfigure}
    \begin{subfigure}[t]{0.32\textwidth}
        \centering
        \includegraphics[width=\textwidth, height=5cm]{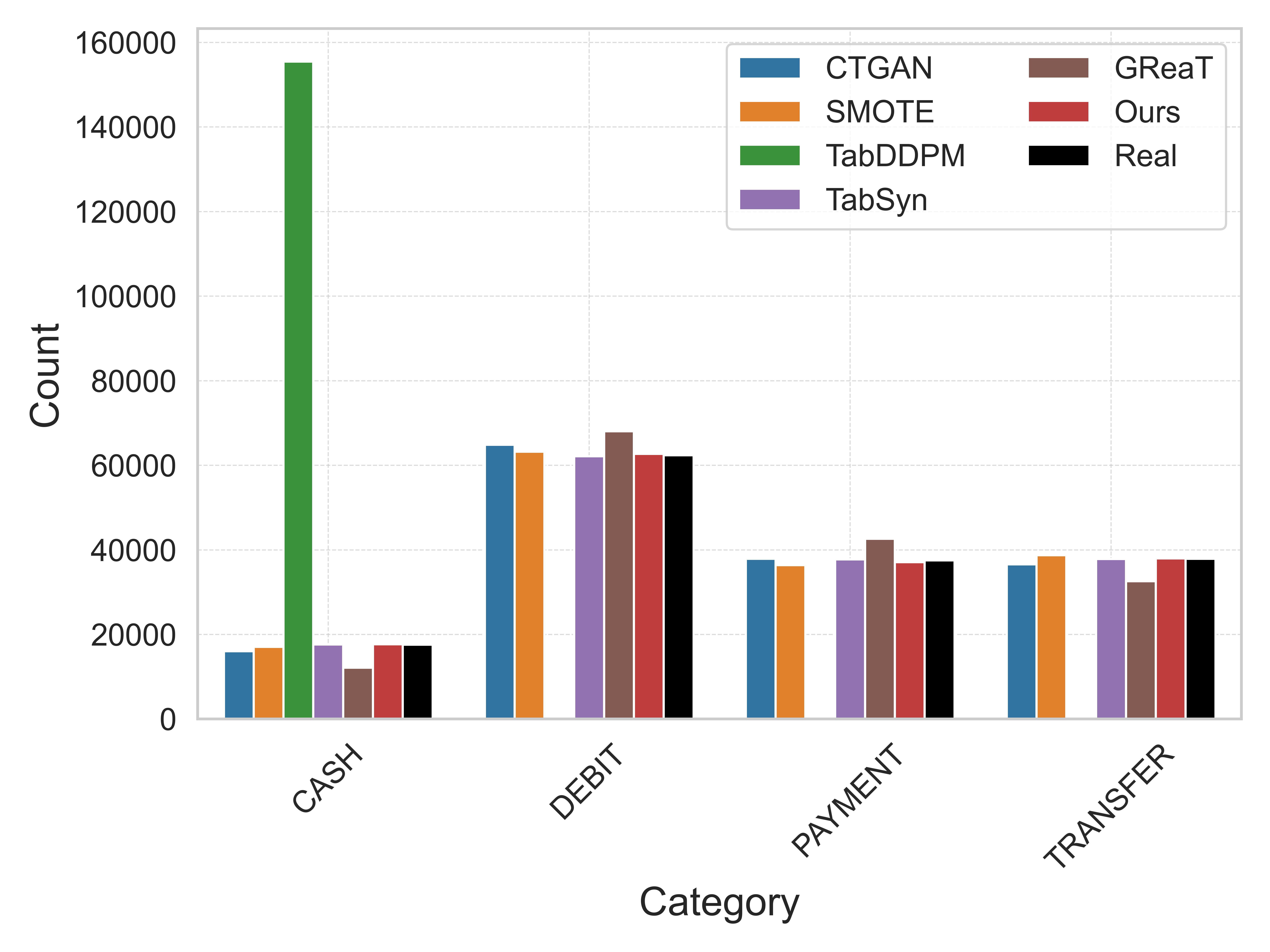}
        \caption{Payment Type}
    \end{subfigure}
    \begin{subfigure}[t]{0.32\textwidth}
        \centering
        \includegraphics[width=\textwidth, height=5cm]{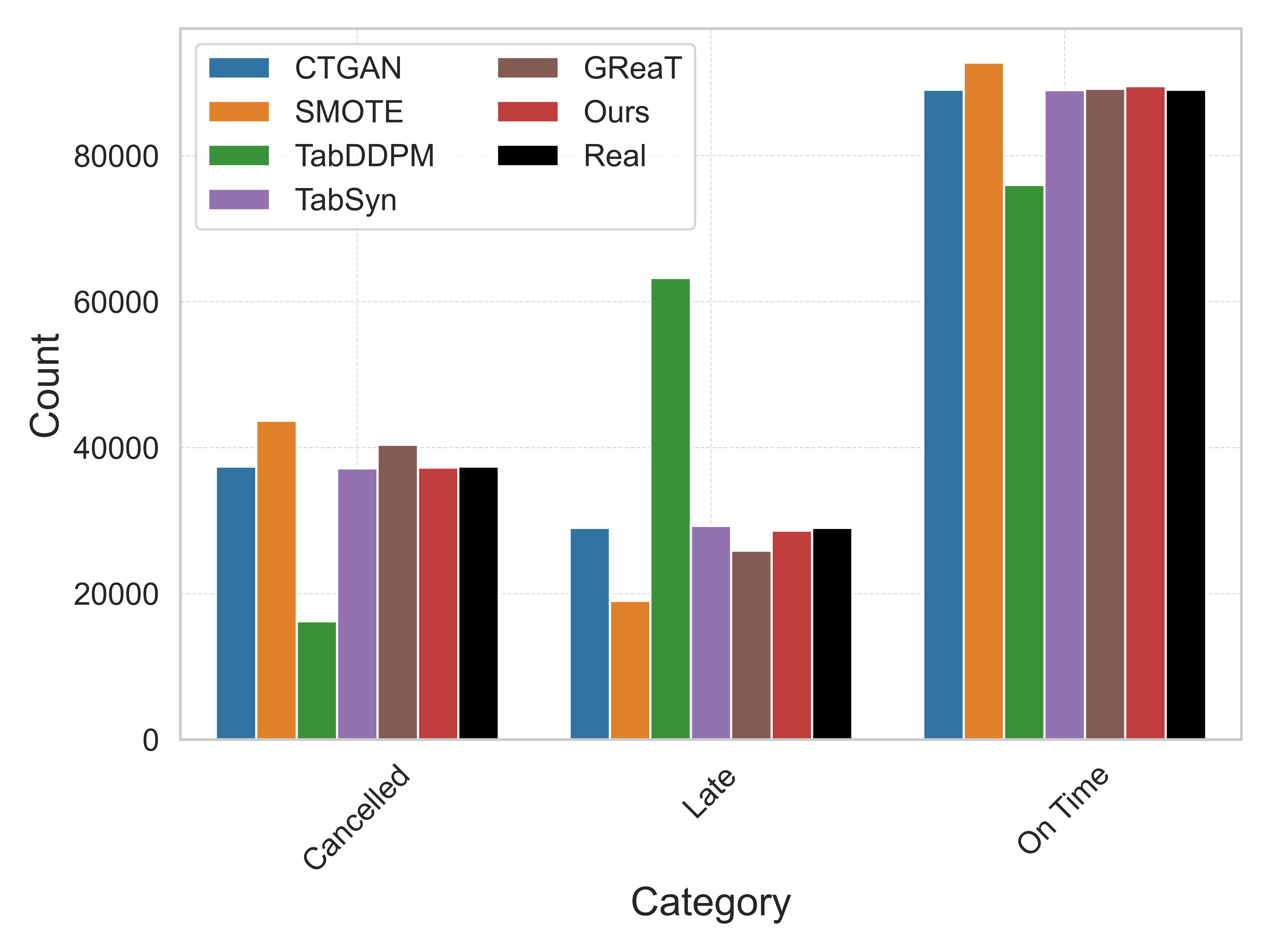}
        \caption{Order Status}
    \end{subfigure}
    \caption{
        Distribution plots for the three {\bf categorical} columns ({\tt shipping mode}, {\tt payment type}, and {\tt order status}), comparing synthetic data generated by CTGAN, TabDDPM, GReaT, TabSyn, and TabKG to real data.
        Distributions that closely match the real data indicate superior performance.
        Both TabKG and TabSyn exhibit distributions that are significantly closer to the real data compared to other methods.
    }
    \label{fig:accuracy2}
\end{figure*}

\begin{figure*}[h!]
    \centering
    \begin{subfigure}[t]{0.32\textwidth}
        \centering
        \includegraphics[width=\textwidth]{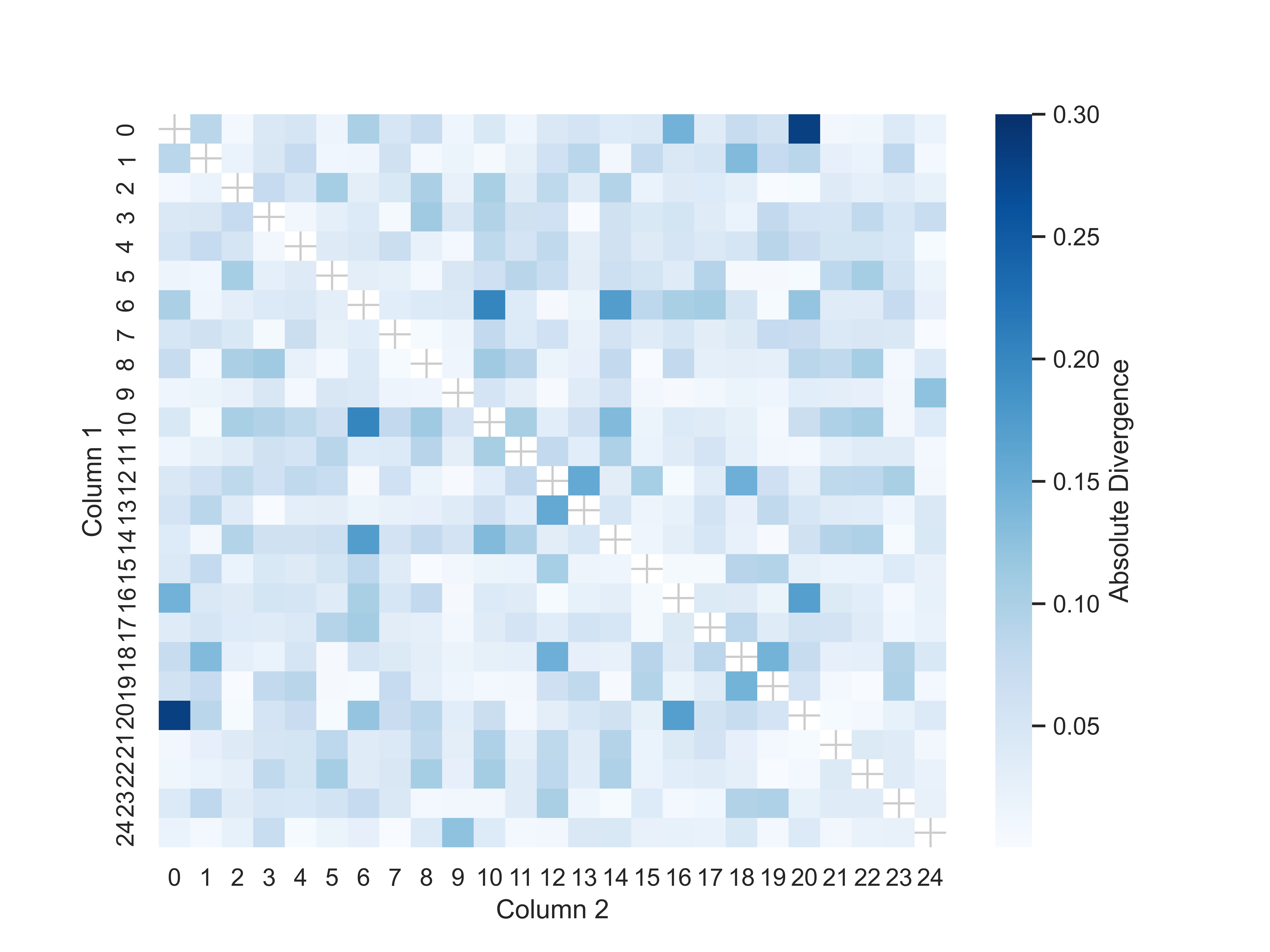}
        \caption{CTGAN}
    \end{subfigure}
    \begin{subfigure}[t]{0.32\textwidth}
        \centering
        \includegraphics[width=\textwidth]{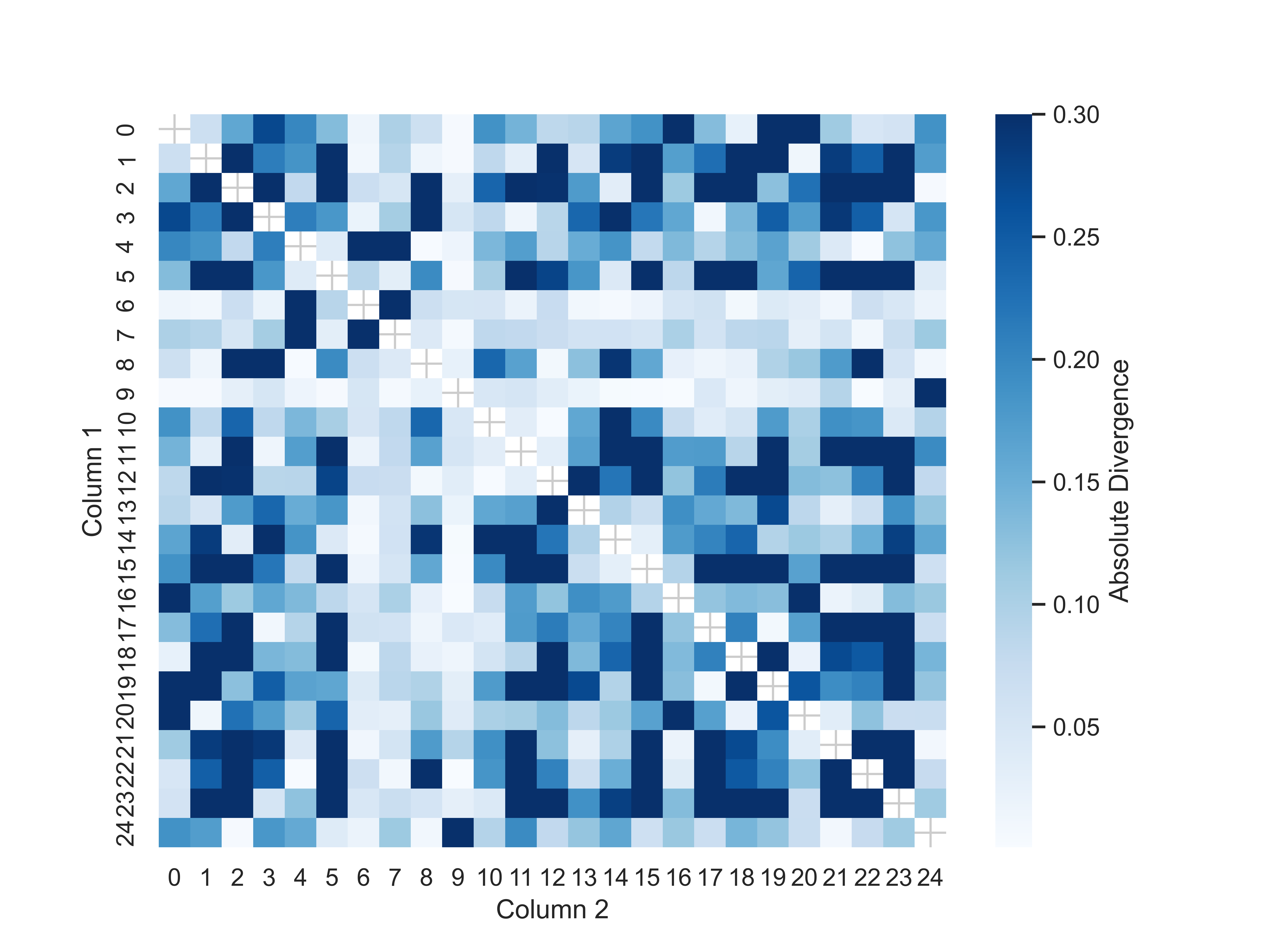}
        \caption{TabDDPM}
    \end{subfigure}
    \begin{subfigure}[t]{0.32\textwidth}
        \centering
        \includegraphics[width=\textwidth]{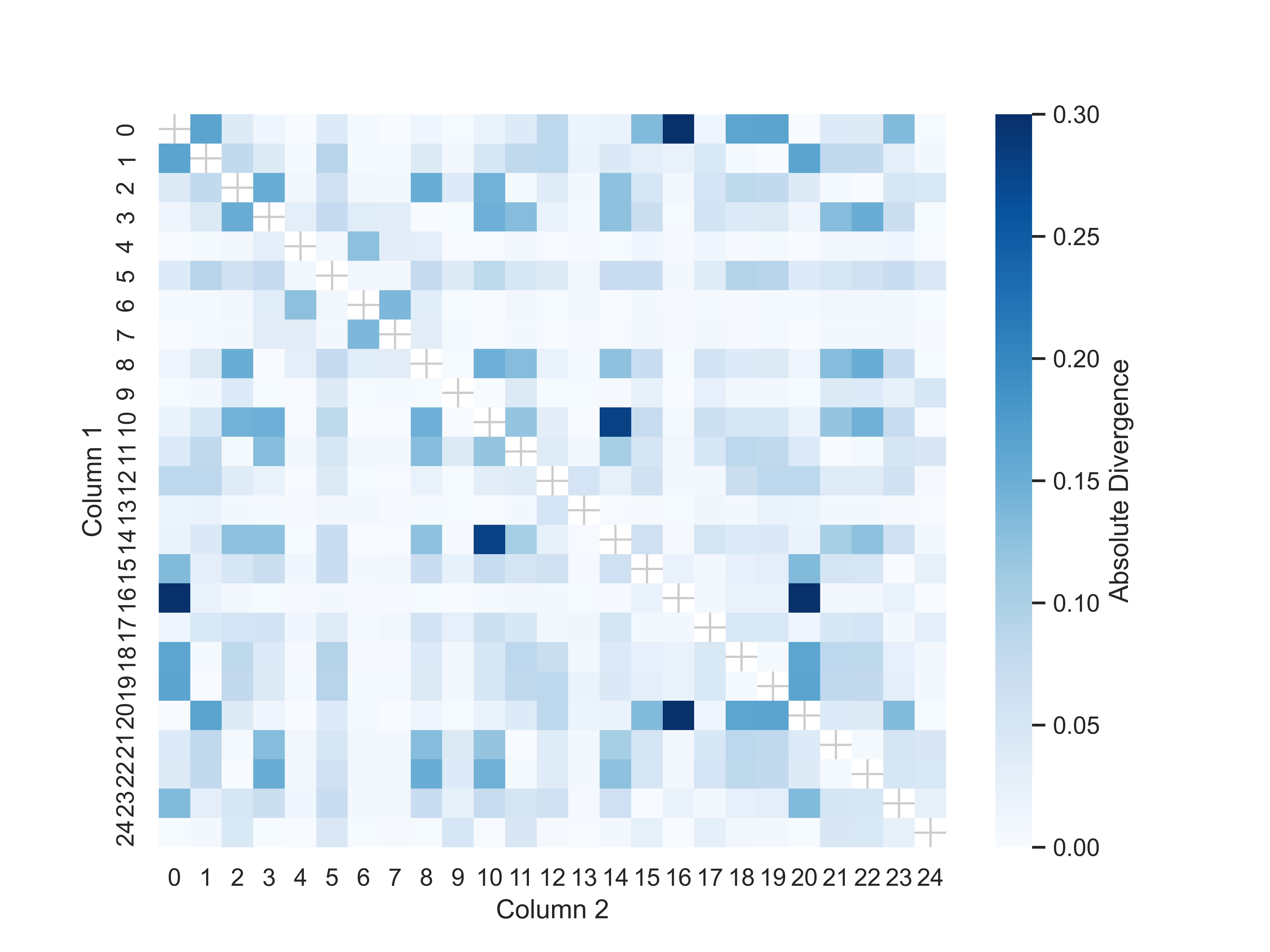}
        \caption{GReaT}
    \end{subfigure}
    \begin{subfigure}[t]{0.32\textwidth}
        \centering
        \includegraphics[width=\textwidth]{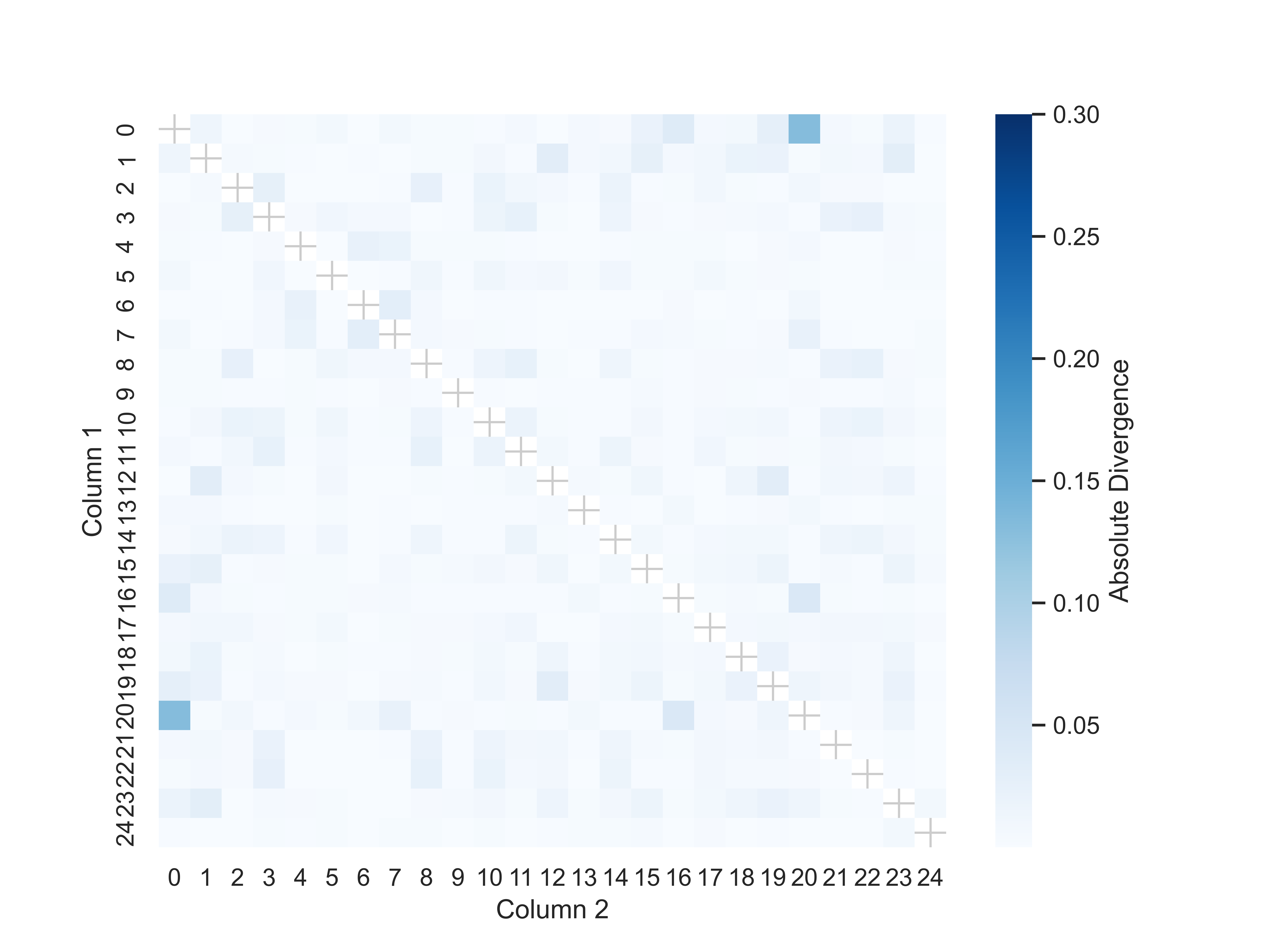}
        \caption{TabSyn}
    \end{subfigure}
    \begin{subfigure}[t]{0.32\textwidth}
        \centering
        \includegraphics[width=\textwidth]{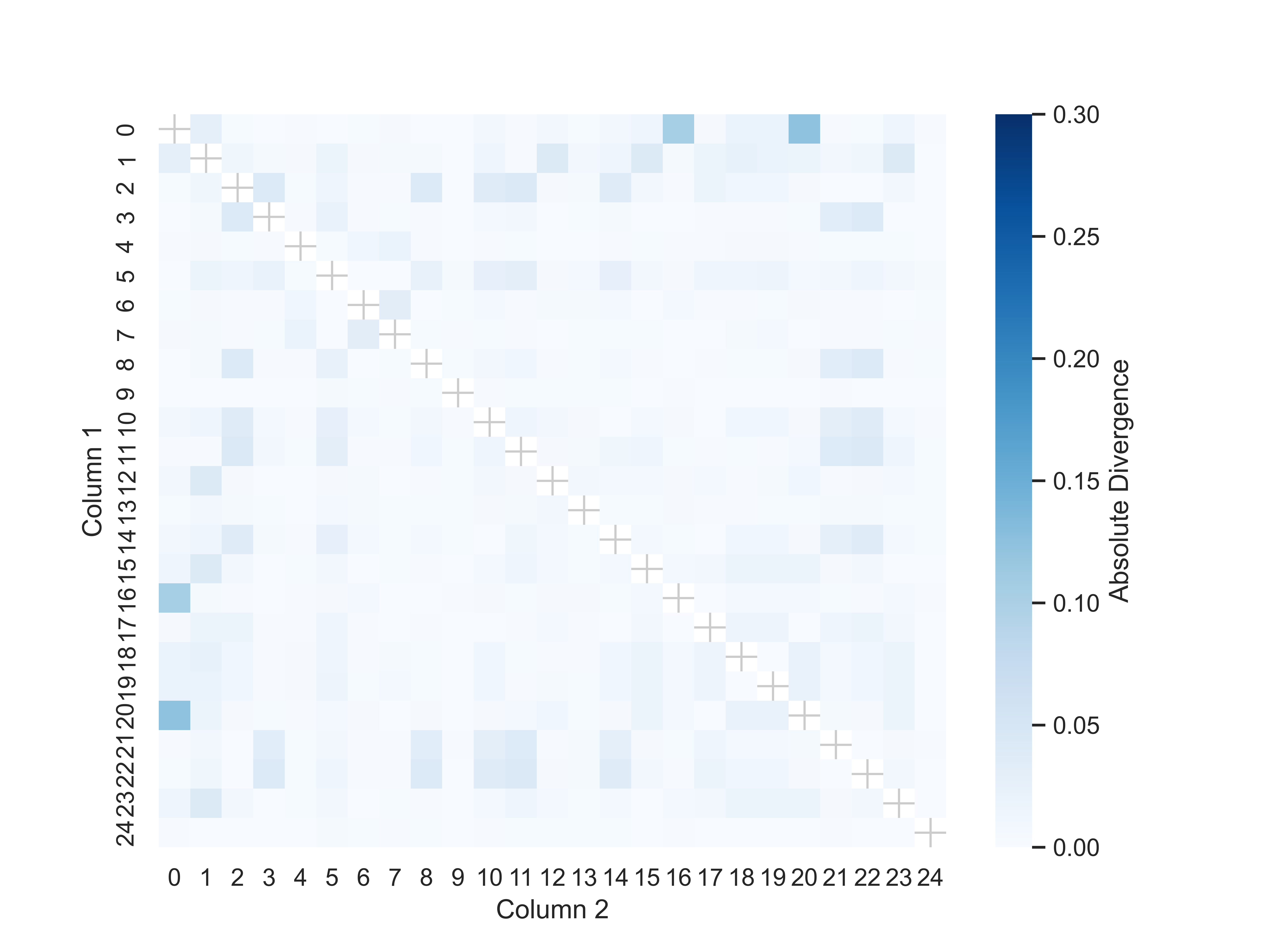}
        \caption{Ours (TabKG)}
    \end{subfigure}
    \caption{Heatmap illustrating the absolute divergence in pairwise column correlations between the synthetic and real data.
    Lighter colors indicate smaller differences and better alignment.
    TabSyn and TabKG exhibit the closest alignment with the real data, outperforming other methods.}
    \label{fig:heatmap}
\end{figure*}

\begin{table}[t!]
\centering
\caption{Inter-column relationships discovered by TabKG on the Retail (36 edges) and Purchasing (14 edges) datasets, grouped by relationship type. The CR-KG construction (Stages 1--3) discovers these dependencies automatically from column metadata, without manual rule engineering.}
\label{tab:relationships}
\renewcommand{\arraystretch}{1.15}
\scriptsize
\begin{tabularx}{\textwidth}{@{}l l X@{}}
\toprule
\textbf{Dataset} & \textbf{Type} & \textbf{Inter-column relationships (selected examples)} \\
\midrule
\multirow{4}{*}{\textbf{Retail}}
& Hierarchical
& \begin{tabular}[t]{@{}l@{}}
\texttt{order\_city} $\to$ \texttt{order\_state} $\to$ \texttt{order\_country} $\to$ \texttt{order\_region} $\to$ \texttt{market} \\
\texttt{customer\_city} $\to$ \texttt{customer\_state} $\to$ \texttt{customer\_country} $\to$ \texttt{customer\_zipcode} \\
\texttt{category\_id} $\to$ \texttt{category\_name}; \quad \texttt{department\_id} $\to$ \texttt{department\_name} \\
\texttt{product\_card\_id} $\to$ \texttt{product\_name}; \quad \texttt{product\_category\_id} $\to$ \texttt{category\_id} \\
\texttt{customer\_id} $\to$ (\texttt{customer\_fname}, \texttt{customer\_lname}, \texttt{customer\_email}, \texttt{customer\_segment})
\end{tabular} \\
\cmidrule(l){2-3}
& Mathematical
& \begin{tabular}[t]{@{}l@{}}
\texttt{sales} = \texttt{order\_item\_quantity} $\times$ \texttt{order\_item\_product\_price} \\
\texttt{order\_item\_discount} = \texttt{sales} $\times$ \texttt{order\_item\_discount\_rate} \\
\texttt{order\_item\_total} = \texttt{sales} $-$ \texttt{order\_item\_discount} \\
\texttt{order\_profit\_per\_order} = \texttt{order\_item\_total} $\times$ \texttt{order\_item\_profit\_ratio}
\end{tabular} \\
\cmidrule(l){2-3}
& Temporal
& \begin{tabular}[t]{@{}l@{}}
\texttt{order\_date} $<$ \texttt{shipping\_date} \\
\texttt{shipping\_date} $-$ \texttt{order\_date} = \texttt{days\_for\_shipping\_real}
\end{tabular} \\
\cmidrule(l){2-3}
& Semantic
& \begin{tabular}[t]{@{}l@{}}
\texttt{late\_delivery\_risk} $\Leftrightarrow$ (\texttt{days\_for\_shipping\_real} $>$ \texttt{days\_for\_shipment\_scheduled}) \\
\texttt{delivery\_status} determined by (\texttt{days\_for\_shipping\_real}, \texttt{days\_for\_shipment\_scheduled}) \\
\texttt{shipping\_mode} $\in$ \{Standard Class, First Class, Second Class, Same Day\} \\
\texttt{type} $\in$ \{DEBIT, TRANSFER, PAYMENT, CASH\}
\end{tabular} \\
\midrule
\multirow{4}{*}{\textbf{Purchasing}}
& Hierarchical
& \begin{tabular}[t]{@{}l@{}}
\texttt{supplier\_number} $\to$ \texttt{supplier\_name} $\to$ \texttt{supplier\_country} \\
\texttt{material\_number} $\to$ \texttt{material\_description} \\
\texttt{plant} $\to$ \texttt{storage\_location}; \quad \texttt{purchase\_organization} $\to$ \texttt{company\_code}
\end{tabular} \\
\cmidrule(l){2-3}
& Mathematical
& \begin{tabular}[t]{@{}l@{}}
\texttt{net\_amount} = \texttt{quantity} $\times$ \texttt{net\_price} \\
\texttt{tax\_amount} = \texttt{net\_amount} $\times$ \texttt{tax\_rate} \\
\texttt{gross\_amount} = \texttt{net\_amount} + \texttt{tax\_amount} \\
\texttt{lead\_time} = \texttt{planned\_delivery\_date} $-$ \texttt{order\_date}
\end{tabular} \\
\cmidrule(l){2-3}
& Temporal
& \texttt{order\_date} $<$ \texttt{planned\_delivery\_date} $<$ \texttt{actual\_delivery\_date} $<$ \texttt{goods\_receipt\_date} \\
\cmidrule(l){2-3}
& Semantic
& \begin{tabular}[t]{@{}l@{}}
\texttt{po\_status} $\in$ \{OPEN, PARTIAL, CLOSED, CANCELED\} \\
\texttt{currency} $\in$ \{USD, EUR, GBP, CNY, $\dots$\} \\
\texttt{document\_type} consistent with \texttt{purchase\_organization}
\end{tabular} \\
\bottomrule
\end{tabularx}
\end{table}

As shown in \autoref{tab:accuracy}, TabKG and TabSyn achieve the best overall accuracy, with TabKG leading on $\alpha$-Precision across both datasets (99.15\% and 97.29\% on the Retail and Purchasing datasets, respectively) and TabSyn leading on pairwise correlation. The density distributions of representative continuous columns (\autoref{fig:accuracy1}) and categorical columns (\autoref{fig:accuracy2}) further demonstrate that TabKG and TabSyn produce distributions nearly identical to the real data, while CTGAN and GReaT exhibit noticeable gaps. The pairwise column correlation heatmaps (\autoref{fig:heatmap}) confirm that TabKG and TabSyn achieve the closest alignment with real data.

For data diversity, TabKG achieves the highest $\beta$-Recall on the Retail dataset (27.85) and the second-highest Average Coverage Score (99.18, marginally below TabSyn at 99.52). On the Purchasing dataset, TabDDPM attains the highest coverage (97.49) and GReaT the highest $\beta$-Recall (36.16), while TabKG remains competitive on both. Overall, TabKG strikes an excellent balance between accuracy and diversity across both datasets.

\subsubsection{Data Fidelity: Inter-Column Relationship Preservation}

\autoref{tab:relationships} reports the inter-column relationships discovered by TabKG, organised by type. The CR-KG construction surfaces a rich set of dependencies on both datasets: full multi-level geographical and product hierarchies, mathematical formulas governing pricing and profit, multi-step temporal orderings between order--shipment events, and categorical (semantic) constraints on status fields. Notably, several discovered relationships go beyond the simple pairwise dependencies typically considered in tabular generation benchmarks; examples include the chain \texttt{order\_city}~$\to$~\texttt{order\_state}~$\to$~\texttt{order\_country}~$\to$~\texttt{order\_region}~$\to$~\texttt{market}, and the conditional rule that \texttt{late\_delivery\_risk} is implied by the comparison of two date columns. These structures are precisely what existing latent-space generators fail to preserve.

\begin{table}[ht]
\centering
\caption{Evaluation results for preserving inter-column relationships in synthetic tabular data. 
Higher values indicate better performance. 
The best and second-best results are highlighted in {\color{blue}\bf blue} and {\bf black boldfaced}, respectively.}
\label{tab:logic}
\renewcommand{\arraystretch}{1.15}
\small
\resizebox{0.85\textwidth}{!}{%
\begin{tabular}{ccccccc}
\toprule
\multirow{2}{*}{\textbf{Datasets}} &
\multirow{2}{*}{\textbf{Metrics}} &
  \multicolumn{3}{c}{\textbf{Latent Space-based}} &
  \textbf{LLM-based} & \textbf{Ours}\\ \cmidrule(lr){3-5} \cmidrule(lr){6-6} \cmidrule(lr){7-7}
 &  &
  \textbf{CTGAN} &
  \textbf{TabDDPM} &
  \textbf{TabSyn} &
  \textbf{GReaT} &
  \textbf{TabKG} \\ \midrule
\multirow{3}{*}{\bf{Retail}} &
{HCS} &
   39.23$\pm$0.13 &
   16.07$\pm$0.21 &
   71.63$\pm$0.28 &
   \textbf{95.62$\pm$0.27} &
   \textcolor{blue}{\textbf{97.84$\pm$0.21}} \\
&
{MDI} &
   38.87$\pm$0.25 &
   59.08$\pm$0.25 &
   68.34$\pm$0.38 &
   \textbf{96.18$\pm$0.24} &
   \textcolor{blue}{\textbf{98.41$\pm$0.18}} \\
&
{DSI} &
   30.55$\pm$0.13 &
   79.72$\pm$0.12 &
    66.94$\pm$0.21 &
   \textbf{97.46$\pm$0.18} &
   \textcolor{blue}{\textbf{97.92$\pm$0.15}} \\\hline
\multirow{3}{*}{\bf{Purchasing}} &
{HCS} &
   0.00$\pm$0.00 &
   0.01$\pm$0.05 &
   54.99$\pm$0.23 &
   \textcolor{blue}{\textbf{98.67$\pm$0.32}} &
   \textbf{98.41$\pm$0.27} \\
&
{MDI} &
   36.69$\pm$0.36 &
   63.49$\pm$0.23 &
   56.30$\pm$0.41 &
   \textbf{94.54$\pm$0.36} &
   \textcolor{blue}{\textbf{98.73$\pm$0.21}} \\
&
{DSI} &
   59.39$\pm$0.23 &
   \textbf{96.62$\pm$0.21} &
   93.15$\pm$0.11 &
   94.45$\pm$0.32 &
   \textcolor{blue}{\textbf{96.83$\pm$0.12}} \\ \bottomrule
\end{tabular}%
}
\end{table}

The quantitative impact of preserving these structures is shown in \autoref{tab:logic}. Averaged over 10 independent generation runs, TabKG attains HCS and MDI scores between 97.84\% and 98.73\% across both datasets, substantially above all latent-space baselines but not exactly 100\%, since edges that are not discovered by the CR-KG (and therefore not enforced deterministically) are left to the diffusion model, allowing a small fraction (roughly 1--2\%) of synthetic records to violate at least one discovered constraint. On the Purchasing dataset, GReaT achieves the highest HCS (98.67\%), marginally above TabKG (98.41\%), owing to its autoregressive modelling which naturally captures sequential dependencies; however, TabKG achieves the highest MDI across both datasets. Latent-space models (CTGAN, TabDDPM, TabSyn) perform significantly worse, particularly in hierarchical consistency: on the Purchasing dataset, CTGAN and TabDDPM collapse to near-zero HCS, and TabSyn reaches only 54.99\%. The DSI scores further indicate that TabKG best preserves implicit distributional dependencies across financial attributes.

\subsubsection{Data Utility on Real Classification Tasks}
\label{sec:utility}

  

\begin{table}[ht]
\centering
\caption{Evaluation results of machine learning ({\bf classification}) efficiency in terms of the AUC score. 
Higher AUC values indicate better performance. 
The best and second-best results are highlighted in {\color{blue}\bf blue} and {\bf black boldfaced}, respectively.}
\label{tab:utility}
\renewcommand{\arraystretch}{1.15}
\small
\resizebox{0.8\textwidth}{!}{%
\begin{tabular}{ccccccc}
\toprule
 &
  \multicolumn{1}{c}{\textbf{Real Dataset}} &
  \multicolumn{3}{c}{\textbf{Latent Space-based}} &
  \multicolumn{1}{c}{\textbf{LLM-based}} &
  \textbf{Ours} \\
  \cmidrule(lr){2-2} \cmidrule(lr){3-5} \cmidrule(lr){6-6} \cmidrule(lr){7-7} 
  \textbf{Datasets} &
  \textbf{Real} &
  \textbf{CTGAN} &
  \textbf{TabDDPM} &
  \textbf{TabSyn} &
  \textbf{GReaT} &
  \textbf{TabKG} \\ \hline
\bf{Retail}  & 88.17 & 68.91±0.18 & 49.68±0.91 & \textbf{71.55±0.24} & 71.15±0.14 & \textcolor{blue}{\textbf{71.90±0.20}} \\
\hline
\bf{Purchasing} & 84.37 & 53.31±0.60 & 49.72±1.28 & \textbf{62.26±1.38} & 50.06±0.75 & \textcolor{blue}{\textbf{63.45±0.92}} \\ \bottomrule
\end{tabular}%
}
\end{table}
To test whether TabKG-generated data is usable for \emph{real operational decision-making}, we evaluate it on two industrial classification tasks drawn from the case studies: \textit{late-delivery risk prediction} on the Retail dataset (a standard SLA-monitoring task used by logistics planners) and \textit{procurement-status classification} on the Purchasing dataset (a task relevant to order-to-cash monitoring). Following the Train-on-Synthetic, Test-on-Real (TSTR) protocol, we train an XGBoost classifier purely on synthetic data generated by each method and evaluate on a held-out split of the \emph{real} operational data; AUC therefore directly reflects how well the synthetic data supports real downstream decisions rather than just distributional fidelity.

As shown in \autoref{tab:utility}, TabKG achieves the highest AUC on both real-world tasks (71.90 on Retail and 63.45 on Purchasing), with TabSyn second-best on both (71.55 and 62.26). GAN-based and diffusion-based baselines (CTGAN, TabDDPM, GReaT) are considerably less effective, with TabDDPM in particular failing to learn useful decision boundaries (AUC near random on Purchasing). All synthetic methods still lag behind models trained on real data (88.17 and 84.37), which sets the practical ceiling; nevertheless, the TabKG-trained classifiers preserve roughly 82\% and 75\% of the real-data AUC, respectively, which is within the range typically considered usable for cross-firm data sharing and SME augmentation scenarios. The margin between TabKG and the non-structure-aware baselines indicates that the inter-column relationship preservation translates directly into operationally meaningful classification performance: logically consistent synthetic data better captures the decision boundaries needed for real supply chain tasks, not just the marginal distributions.

\subsubsection{Data Privacy Preservation}\label{sec:privacy}

\begin{table}[h]
\centering
\caption{Evaluation results of privacy preservation in terms of the DCR and C2ST. 
Lower DCR values indicate better privacy preservation, with the best and smallest value being 50\%.
Higher C2ST values indicate better performance in terms of accuracy. 
The best and second-best results are highlighted in {\color{blue}\bf blue} and {\bf black boldfaced}, respectively.}
\label{tab:privacy}
\renewcommand{\arraystretch}{1.15}
\small
\resizebox{0.85\textwidth}{!}{%
\begin{tabular}{ccccccc}
\toprule[1pt]
\multirow{2}{*}{\textbf{Datasets}} &
\multirow{2}{*}{\textbf{Metrics}} &
  \multicolumn{3}{c}{\textbf{Latent Space-based}} &
  \textbf{LLM-based} & \textbf{Ours}\\ \cmidrule(lr){3-5} \cmidrule(lr){6-6} \cmidrule(lr){7-7}
 &  &
  \textbf{CTGAN} &
  \textbf{TabDDPM} &
  \textbf{TabSyn} &
  \textbf{GReaT} &
  \textbf{TabKG} \\ \hline
\multirow{2}{*}{\textbf{Retail}} &
DCR & \textbf{89.80±0.36} & \textcolor{blue}{\textbf{89.15±0.28}} & 90.18±0.27 & 90.19±0.25 & 90.03±0.23 \\
&
    C2ST & 17.65±0.32 & 0.00±0.00 & \textbf{52.89±0.23} & 38.83±0.16 & \textcolor{blue}{\textbf{66.75±0.15}} \\ \hline
\multirow{2}{*}{\textbf{Purchasing}} &
DCR & 90.67±0.20 & 95.02±0.21 & \textbf{90.60±0.24} & 91.89±0.25 & \textcolor{blue}{\textbf{90.43±0.22}} \\ 
&
    C2ST & 66.92±0.69 & 0.27±0.13 & \textbf{93.93±0.39} & 76.34±0.37 & \textcolor{blue}{\textbf{95.38±0.30}} \\ \bottomrule
\end{tabular}%
}
\end{table}

As shown in \autoref{tab:privacy}, TabKG achieves the best balance between privacy (DCR) and accuracy (C2ST) across both datasets. While TabDDPM achieves competitive DCR values, its extremely low C2ST indicates that privacy comes at the cost of data quality. TabKG provides both robust privacy protection and high data fidelity, making it well-suited for industrial scenarios requiring both guarantees.

\subsubsection{Computational Cost}\label{sec:cost}
\autoref{tab:cost} reports the approximate wall-clock times for each stage of the TabKG pipeline and for the baseline generative models on the Retail dataset. The CR-KG construction (Stages 1--3) requires approximately 5 minutes when using a five-model ensemble, dominated by LLM API inference time. The compression and decompression stages are negligible ($<$1 second). The latent diffusion training (Stage 4) takes approximately 30 minutes on a single NVIDIA RTX 4090 GPU, comparable to TabSyn since the same diffusion architecture is used on a reduced column set. Among the baselines, GReaT requires the longest training time due to its autoregressive LLM fine-tuning, while CTGAN is the fastest. TabKG's total pipeline time is competitive with TabSyn while providing substantially stronger logical consistency guarantees.

\begin{table}[h]
\centering
\caption{Computational efficiency comparison on the Retail dataset.}
\label{tab:cost}
\renewcommand{\arraystretch}{1.15}
\resizebox{0.7\textwidth}{!}{%
\begin{tabular}{lcc}
\toprule
\textbf{Method / Stage} & \textbf{Training} & \textbf{Sampling} \\
\midrule
CTGAN & $\sim$10 min & $<$1 min \\
TabDDPM & $\sim$45 min & $\sim$5 min \\
TabSyn & $\sim$30 min & $\sim$2 min \\
GReaT & $\sim$5 hours & $\sim$30 min \\
\midrule
TabKG: CR-KG construction (Stages 1--3) & $\sim$5 min & --- \\
TabKG: Diffusion training (Stage 4) & $\sim$30 min & $\sim$2 min \\
TabKG: Decompression (Stage 5) & --- & $<$1 sec \\
\textbf{TabKG total} & $\sim$\textbf{35 min} & $\sim$\textbf{2 min} \\
\bottomrule
\end{tabular}%
}
\end{table}

\section{Results Discussion}

The results show that TabKG is most effective when the target table contains explicit operational structure. In the Purchasing dataset, the schema is compact, with 21 columns and 14 ground-truth relationships, and TabKG achieves an F1 score of 0.97 for relationship discovery. It correctly recovers the procurement-cycle temporal ordering, \texttt{order\_date} $<$ \texttt{planned\_delivery} $<$ \texttt{actual\_delivery} $<$ \texttt{goods\_receipt}, as well as the net/tax/gross mathematical chain. These are precisely the types of relationships needed for downstream simulators of supplier lead-time variability, delivery reliability, and procurement working-capital flows.
The Retail dataset presents a harder setting. It is wider and noisier, with 41 columns and 36 ground-truth relationships, including ambiguous customer attributes and indirect label-derived relationships. TabKG still achieves an F1 score of 0.90, but the missed edges concentrate in relationships that cannot be reliably inferred from column metadata alone, such as \texttt{customer\_segment} inferred from a combination of customer attributes. This suggests that the practical value of TabKG depends on the schema in scope. ERP- and SCM-style tables with well-structured identifiers, timestamps, financial fields, and transactional attributes benefit most directly, whereas customer-facing tables with implicit or behavioural relationships may require additional documentation, data samples, or domain-specific metadata before LLM-based reasoning can recover all dependencies.

The comparison with alternative generator families further clarifies TabKG's positioning. GReaT, the autoregressive LLM-based generator, comes closest in hierarchical consistency, achieving HCS = 98.67\% compared with TabKG's 98.41\% on the Purchasing dataset. However, it does so at substantially higher training and sampling cost, and its outputs occasionally contain category values absent from the training vocabulary. This is problematic for ERP-grade simulation, where categorical domains such as supplier identifiers, product codes, or procurement statuses must remain valid. In contrast, latent-space generators such as CTGAN, TabDDPM, and TabSyn are computationally efficient, but they fail to preserve key hierarchical chains: on the Purchasing dataset, CTGAN and TabDDPM collapse to near-zero HCS, and TabSyn reaches only 54.99\%, well below the threshold required for ERP-grade simulation.

Overall, these findings support the central claim that trustworthy synthetic supply chain data requires more than statistical realism. A generator can match marginal distributions and downstream utility while still violating temporal, mathematical, or hierarchical constraints that govern real operations. By representing such constraints in a validated CR-KG and using them to guide generation, TabKG moves synthetic tabular data closer to operationally valid simulation and decision-support use cases.

\subsection{Managerial Implications}

The broader managerial significance of TabKG lies in enabling AI systems to understand and reason about supply chains as they actually operate in the real physical world. By generating synthetic data that respects the operational ``physics'' of supply chain processes, including temporal orderings, mathematical formulas, hierarchical taxonomies, and conditional rules, TabKG provides a grounded substrate on which AI-driven decision support, simulation, and autonomous orchestration in \textit{Supply Chain 4.0} can be built without drifting from the operational reality on the ground. We organise the managerial insights around three key supply chain challenges where this grounding has direct practical impact.

\subsubsection{Privacy-preserving synthetic data for supply chain machine learning}
A central practical use of synthetic data in supply chains is to support machine learning tasks, such as demand forecasting, predictive maintenance, supply chain risk prediction, and payment-delay or fraud detection, that are routinely limited by privacy concerns over sensitive operational records and by data scarcity, particularly among SMEs with limited historical data \citep{long2025leveraging}. TabKG addresses both barriers simultaneously. First, it preserves \emph{privacy}: by training only on the compressed independent-column subset and reconstructing dependent columns deterministically, TabKG produces synthetic records that are statistically faithful yet do not reproduce raw sensitive entries (Table~\ref{tab:privacy}), enabling organisations to use realistic datasets without exposing proprietary information. Second, it preserves \emph{operational utility}: because the CR-KG enforces inter-column logical relationships, the synthetic data can be used to augment the performance for downstream supply chain ML tasks where logical consistency is required for the model to learn meaningful decision boundaries \citep{wyrembek2025causal, lokanan2025supply, kong2026hierarchical, schneckenreither2021order}.

\subsubsection{Enabling high-fidelity supply chain simulation}
Data-driven simulation architectures \citep{luo2022data, khayyati2022machine, xu2025multi, polini2020digital} require input data that respects domain constraints; logically inconsistent synthetic data would otherwise produce unrealistic and misleading results. TabKG enables a new simulation paradigm: generating unlimited synthetic scenarios where every order has valid pricing, consistent geography, and feasible delivery timelines. This allows practitioners to stress-test supply chain networks under realistic disruption scenarios (e.g., supplier failures, demand surges, logistics bottlenecks) and to validate agent-based simulations \citep{xu2024multi} with logically consistent data, ensuring that operational decisions derived from simulations are grounded in reality.

\subsubsection{Supporting LLM-driven reasoning and decision-making in supply chains}
As foundation models are increasingly applied to supply chain decision-making \citep{xu2024multi}, they require both large volumes of domain-grounded training data and explicit representations of operational logic. TabKG's CR-KG serves a dual purpose: (1) generating logically consistent training data at scale to teach foundation models the ``physics'' of supply chain operations, including temporal constraints, pricing formulas, and hierarchical structures; and (2) providing a structured knowledge base that LLMs can query during inference to reason about domain-specific constraints. This bridges the gap between general-purpose language models and the domain-specific logic required for autonomous supply chain management, enabling more reliable LLM-driven planning, procurement optimisation, and disruption response.

\subsubsection{Leveraging pre-trained LLM knowledge for adaptive synthetic data generation across industries}
Conventional tabular generators learn their underlying logic solely from the training data, which presumes the data is clean, representative, and sufficiently rich to expose the relevant operational rules. In practice, real supply chain data are often noisy, sparse, or skewed, particularly among SMEs and in emerging markets, which severely limits what a data-only generator can learn from them and can introduce significant drawbacks for downstream applications \citep{long2025leveraging}. TabKG offers a complementary path: rather than relying solely on the training data, it leverages the pre-trained knowledge already encoded in LLMs to surface candidate inter-column operational logic directly from column metadata, before any generative training begins. Two managerial consequences follow. First, the CR-KG construction is \emph{training-free}: no labelled rules, manual ontology engineering, or in-the-loop domain expert is needed to guide the downstream diffusion model toward high-fidelity, logically consistent output. Second, because the LLM-derived logic does not depend on any single industry's hard-coded rule set, the same TabKG pipeline can be applied across diverse tabular schemas spanning retail, procurement, manufacturing, and logistics without bespoke rule libraries. This lowers the barrier to deployment for firms that cannot afford a dedicated knowledge-engineering function, and makes logically consistent synthetic data accessible to a much broader range of supply chain organisations.

In practical deployment, two readiness considerations follow from this design. First, because TabKG's reasoning quality is anchored in LLM interpretation of column metadata, firms that maintain reasonably descriptive column names and accompanying schema documentation will see the largest benefits, while organisations operating on legacy ERP systems with cryptic identifiers (e.g., \texttt{LIFNR}, \texttt{MATNR}) should treat lightweight metadata curation, namely the addition of a one-line natural-language description per column, as a low-cost preparatory step that unlocks the rest of the pipeline. Second, the training-free, schema-agnostic property means TabKG can be redeployed quickly when schemas change (e.g., after an ERP migration or the introduction of a new product line), avoiding the rebuilding cost typically associated with rule-based or schema-specific generators. Together, these properties make TabKG well suited to phased adoption: an organisation can pilot it on a single well-documented table, validate the discovered operational logic with domain experts, and then scale to additional tables and business units as metadata is incrementally curated, rather than requiring an upfront firm-wide knowledge-engineering investment.

\section{Limitations}

A key limitation for industrial supply chain deployment is the quality of input column metadata. TabKG's reasoning quality depends on column names being reasonably descriptive and, ideally, accompanied by clear textual descriptions. Many legacy ERP schemas contain cryptic identifiers, such as \texttt{LIFNR} or \texttt{MATNR}, missing descriptions, or organization-specific abbreviations. In such settings, TabKG's relationship-discovery performance may degrade unless the metadata is curated, augmented with documentation, or adapted to in-house naming conventions.

A second limitation concerns data standardization. Real operational datasets often mix heterogeneous units, currencies, locales, and time zones across regions, suppliers, and business units. The current CR-KG construction process does not explicitly normalize these differences before inferring relationships. As a result, some valid mathematical or temporal dependencies may be missed or incorrectly specified when equivalent fields are recorded using inconsistent units or regional conventions.

A third limitation is that our experiments focus on normal-regime operational data. Supply chains often shift between stable operating regimes and disruption regimes, such as stockouts, supplier failures, demand shocks, port congestion, or emergency substitutions. Our evaluation does not yet test whether the discovered logical structure remains stable under such distribution shifts. Although many operational constraints, such as temporal ordering or tax calculations, should remain valid across regimes, empirical relationships learned from normal operations may not transfer cleanly to disrupted conditions.

Finally, TabKG currently evaluates logical consistency through predefined relationship categories, including hierarchical, mathematical, temporal, and semantic dependencies. These categories cover common supply chain tabular structures, but they may not capture all forms of operational logic, such as causal dependencies, policy constraints, or exception-handling rules. Extending CR-KG construction to richer rule types is therefore an important direction for future work.

\section{Conclusion and Future Work}
\label{sec:conclusion}

This paper has addressed a fundamental challenge in production research: the generation of synthetic tabular data that respects the inter-column logical structure of supply chain operations. We proposed TabKG, a novel approach that integrates LLM-driven reasoning with latent score-based diffusion. To evaluate this framework, we introduced a multi-dimensional assessment suite focusing on fidelity, utility, privacy and, crucially, logical consistency. Our experimental results across diverse industrial datasets demonstrate that TabKG effectively maintains hierarchical dependencies and business rules where traditional GAN and diffusion-based methods fail, all while providing robust privacy protection.
To the best of our knowledge, TabKG is the first framework to explicitly preserve complex inter-column relationships in supply chain tabular data without manual rule engineering, transforming synthetic data from a machine learning augmentation tool into a reliable resource for industrial simulation and advanced decision support.
Future work will pursue three directions: (i) domain-specific fine-tuning of LLMs to handle abbreviated or noisy column headers common in legacy industrial systems, and integration with real-time streaming pipelines for live simulation applications \citep{afif2025computer}; (ii) combining TabKG with causal inference \citep{wyrembek2025causal} to generate not only logically consistent but also causally faithful data for what-if analysis; and (iii) broader evaluation across automotive, pharmaceutical, and semiconductor manufacturing contexts to assess generalisability \citep{cheng2022linkages}.

\section*{Data Availability Statement}

The datasets used in this study include both public and restricted-access sources.
The DataCo Supply Chain Dataset~\citep{dataco} is publicly available.
The Purchasing Dataset is proprietary and cannot be shared due to confidentiality agreements.

\bibliographystyle{iclr2025_conference}
\bibliography{reference}

\end{document}